\documentclass[lettersize,journal]{IEEEtran}
\usepackage{float} 
\usepackage{amsmath,amsfonts}
\usepackage{algorithmic}
\usepackage{algorithm}
\usepackage{array}
\usepackage[caption=false,font=normalsize]{subfig}
\usepackage{textcomp}
\usepackage{stfloats}
\usepackage{url}
\usepackage{verbatim}
\usepackage{graphicx}
\usepackage{cite}
\usepackage{multirow}
\usepackage{tabularx,booktabs}
\newcolumntype{C}{>{\centering\arraybackslash}X} 
\usepackage[dvipsnames]{xcolor}
\usepackage{longtable}
\usepackage{makecell, cellspace, caption}
\usepackage{array}
\usepackage{subcaption}
\usepackage{cuted}
\usepackage{multirow}
\usepackage[utf8]{inputenc}
\usepackage{tabularx}
\usepackage{array}
\usepackage{amssymb}
\usepackage{makecell}
\usepackage[margin=2cm]{geometry}
\usepackage{adjustbox}
\usepackage{multirow}
\usepackage{tabulary,booktabs}
\usepackage{ragged2e}
\usepackage{amsmath}
\usepackage{booktabs}
\usepackage{array}
\newcolumntype{L}{>{}l<{}}
\newcolumntype{C}{>{}c<{}}
\newcolumntype{R}{>{}r<{}}
\newcolumntype{P}{>{}p{3.5em}<{}}
\newcolumntype{Z}{>{}p{4em}<{}}
\newcolumntype{F}{>{}p{3.1em}<{}}
\newcolumntype{A}{>{}p{2.5em}<{}}
\newcolumntype{B}{>{}p{3em}<{}}
\newcolumntype{D}{>{}p{2em}<{}}
\newcolumntype{E}{>{}p{1.5em}<{}}
\newcolumntype{G}{>{}p{1em}<{}}
\newcolumntype{H}{>{}p{0.6cm}<{}}
\newcolumntype{I}{>{}p{0.4cm}<{}}
\newcolumntype{J}{>{}p{0.3cm}<{}}

\usepackage{amssymb}
\usepackage{pifont}
\usepackage{graphicx}
\usepackage{hyperref}
\hypersetup{colorlinks,allcolors=black}
\usepackage{flushend}
\usepackage{mathtools}
\begin{document}

\title{$\ell_0$-Regularized Sparse Coding-based Interpretable Network for Multi-Modal Image Fusion}

\author{Gargi Panda, Soumitra Kundu, Saumik Bhattacharya, Aurobinda Routray, \IEEEmembership{Member, IEEE}
\thanks{Gargi Panda and Aurobinda Routray are with the Department of EE, IIT Kharagpur, India
(email: pandagargi@gmail.com; aroutray@ee.iitkgp.ac.in)}
\thanks{Soumitra Kundu is with the Rekhi Centre of Excellence for the Science of Happiness, IIT Kharagpur, India (e-mail: soumitra2012.kbc@gmail.com).}
\thanks{Saumik Bhattacharya is with the Department of E\&ECE, IIT Kharagpur, India
(email: saumik@ece.iitkgp.ac.in)}
}

\markboth{\LaTeX\ Journal }%
{Shell \MakeLowercase{\textit{et al.}}: A Sample Article Using IEEEtran.cls for IEEE Journals}


\maketitle
\begin{abstract}
Multi-modal image fusion (MMIF) enhances the information content of the fused image by combining the unique as well as common features obtained from different modality sensor images, improving visualization, object detection, and many more tasks. In this work, we introduce an interpretable network for the MMIF task, named FNet, based on an $\ell_0$-regularized multi-modal convolutional sparse coding (MCSC) model. Specifically, for solving the $\ell_0$-regularized
CSC problem, we design a learnable $\ell_0$-regularized
sparse coding (LZSC) block in a principled manner through deep unfolding. Given different modality source images, FNet first separates the unique and common features from them using the LZSC block and then these features are combined to generate the final fused image. Additionally, we propose an $\ell_0$-regularized MCSC model for the inverse fusion process. Based on this model, we introduce an interpretable inverse fusion network named IFNet, which is utilized during FNet's training. Extensive experiments show that FNet achieves high-quality fusion results across eight different MMIF datasets. Furthermore, we show that FNet enhances downstream object detection \textcolor[rgb]{ 0,  0,  0}{and semantic segmentation} in visible-thermal image pairs.  We have also visualized the intermediate results of FNet, which demonstrates the good interpretability of our network. Link for code and models: \url{https://github.com/gargi884/FNet-MMIF}.
\end{abstract}

\section{Introduction}
\IEEEPARstart{M}{ulti}-modal image fusion (MMIF) is an active area of research for many years. In MMIF, the aim is to integrate information captured using different modality sensor images into a single fused image, offering enhanced information content compared to individual single-modality sensor images. This improves visualization and interpretation, making the fused image more appropriate for downstream tasks like object detection, segmentation, disease
diagnosis, and more \cite{SwinFusion,coconet}. 

\textcolor[rgb]{ 0,  0,  0}{A common approach to MMIF involves independently extracting features from each source image and subsequently combining them to reconstruct the fused image. Traditional fusion methods employed multi-scale transforms \cite{laplacian} or sparse representations \cite{mdlatlrr} to learn features separately from the source images. However, in an alternative MMIF approach, some methods \cite{coupled,icassp} use dictionary learning to explicitly extract both common and unique features from the source images, which are then combined to generate the fused image. As different modality sensors capture images of the same scene, they inherently contain shared (common) features as well as modality-specific (unique) features. The explicit extraction of both types of features facilitates a more effective integration of complementary information across modalities \cite{icassp}. Though the traditional methods have good interpretability of the underlying fusion process, they depend on time-consuming optimization process and hand-crafted fusion rules, which limit their applicability for large-scale data. The emergence of deep learning (DL)- based methods \cite{auifnet, U2fusion, SwinFusion, murf, laph, cddfuse, mda, itfuse, crossfuse,emma} has made significant progress in image fusion. With a good choice of network architecture and training strategy, these methods can achieve superior performance than the traditional methods. However, pure DL-based methods typically employ empirical, trial-and test strategies to construct the network architecture, which makes the underlying fusion process difficult to interpret and understand.} 

To improve interpretability, researchers have explored algorithm unrolling-based models \cite{lrrnet, cunet}. These models are designed by unrolling traditional MMIF optimization algorithms into deep neural network (DNN) architectures. For example, LRRNet \cite{lrrnet} decomposes each modality source image independently into sparse and low-rank components by unrolling a low rank representation-based optimization method. However, during decomposition, LRRNet does not exploit cross-modal dependencies. CUNet \cite{cunet}, another optimization-inspired method, uses an $\ell_1$-regularized multi-modal convolutional sparse coding (MCSC) model to separate the source images into unique and common features. This separation improves fusion quality by incorporating the shared information across modalities, making it more effective than decomposing the source images separately \cite{icassp}. CUNet leverages learned convolutional sparse coding (LCSC) blocks \cite{lcsc} to extract both unique and common features from multi-modal inputs. These LCSC blocks are implemented by unrolling a convolutional extension of the iterative shrinkage thresholding algorithm (ISTA) \cite{ista} to estimate $\ell_1$-regularized sparse features. The ISTA framework employs a soft thresholding function that nullifies features whose absolute values fall below a predetermined threshold. This mechanism indeed promotes sparsity by discarding negligible features. However, it does so at a cost: in addition to eliminating small features, the soft thresholding operation also reduces the magnitude of the larger, more significant features. As a result, the overall estimation becomes biased, with the effective magnitude of the extracted features being lower than what would be achievable with an optimal sparse solution \cite{weightedl1l0,glista}. This issue is particularly critical in multi-modal image fusion (MMIF) where the goal is to accurately select and integrate the most salient features important for the fusion task. \textcolor[rgb]{ 0,  0,  0}{The uniform shrinkage imposed by $\ell_1$-regularization may lead to fused images with reduced contrast \cite{preserve_edge,tv_handbook,preserve_contrast,l0_tip}}, thereby affecting the quality of the fusion process. However, $\ell_0$-regularization directly penalizes the number of nonzero features without shrinking the remaining values. In other words, while $\ell_0$-regularization still nullifies features with absolute values below a threshold, it preserves the magnitudes of the prominent features, offering an optimal solution of the underlying sparse representation \cite{lcsc_analysis}. Hence, replacing the $\ell_1$-regularization constraint on the unique and common features with an $\ell_0$-based approach results in improved feature estimation, which in turn enhances the design of interpretable networks for MMIF tasks. \textcolor[rgb]{ 0,  0,  0}{A more in-depth comparison of these two regularization techniques, and and an experiment to show the effectiveness of $\ell_0$-regularization in MMIF task are provided in Section I and II of the supplementary material, respectively.}

Moreover, for the MMIF task, where the ground truth fused image is not available, considering an inverse fusion process proves effective \cite{sdnet,emma}. In the inverse fusion process, the fused image is separated back into its source images, and as the quality of these separated images is dependent on the quality of the fused image, constraining them to be similar to the original source images helps enforce the generation of a higher-quality fused image. \textcolor[rgb]{ 0,  0,  0}{Given these limitations, it becomes essential to design an interpretable MMIF network that effectively captures cross-modal dependencies and preserves salient features without losing structural details. To avoid bias from $\ell_1$-regularization, the network may employ a sparsity constraint that retains true feature magnitudes while suppressing noise. Additionally, in the absence of ground truth fused images, the framework may include an inverse fusion process to guide training and ensure high-quality fusion.}

\textcolor[rgb]{ 0,  0,  0}{To address these challenges, we propose FNet—an algorithm unrolling-based interpretable DNN for the MMIF task. Unlike LRRNet, FNet explicitly decomposes inputs into unique and common features while effectively modeling inter-modal correlations. Rather than using $\ell_1$-regularization like CUNet, which can suppress feature magnitudes and introduce bias, FNet adopts an $\ell_0$-regularized MCSC model to preserve optimal salient features. Furthermore, an inverse fusion network, IFNet is integrated to guide the training process and enhance fusion quality. Our novel $\ell_0$-regularized MCSC model, represent each modality source image as a combination of $\ell_0$ -regularized unique and common sparse features.} Based on this model, we develop FNet to solve the MMIF task. Due to such a model-based design, FNet has the benefits of both model-based and DNN. Model-based methods promote prior domain knowledge and interpretability of the underlying physical process, whereas DL methods are very efficient in learning from large-scale data \cite{algorithm}. 
Since no existing work has designed a DNN to solve the $\ell_0$-regularized CSC problem, we propose a novel learnable $\ell_0$-regularized sparse coding (LZSC) block. To design the LZSC block, we first develop an iterative algorithm for solving the $\ell_0$-regularized CSC problem and then unroll this algorithm into our learnable LZSC block. 
In FNet, first, the unique and common features are separated from different modality source images using our proposed LZSC blocks. Then, these features are integrated to get the final fused image. 
In our work, constraining the unique and common features to be $\ell_0$-regularized solves the fundamental formulation of sparse coding and estimates accurate sparse features. Moreover, we propose a novel MCSC model to represent the inverse fusion process. In this model, we represent the fused image as a combination of $\ell_0$-regularized sparse features corresponding to the different modality source images. Based on this model, we design an interpretable inverse fusion network named IFNet. Given the fused image, IFNet first estimates the features corresponding to the source images using our LZSC blocks and then obtains the source images using the convolution operation. We utilize IFNet in the training of FNet. Constraining the decomposed source images obtained by IFNet to be similar to the original source images improves the scene representation quality in the fused image.
Our primary contributions is outlined below:
\begin{enumerate}
\item We design a learnable LZSC block \textcolor[rgb]{ 0,  0,  0}{in a principled manner through deep unfolding} to solve the $\ell_0$-regularized CSC problem. To the best of our knowledge, this is the first learnable block to solve the $\ell_0$-regularized CSC problem.
\item We introduce an MCSC model to address the MMIF task. In this model, we represent each modality source image as a combination of $\ell_0$-regularized unique and common sparse features. Based on this model and our LZSC block, we propose an interpretable fusion network named FNet.
\item We introduce an MCSC model for the inverse fusion process. In this model, we represent the fused image as a combination of $\ell_0$-regularized sparse features corresponding to the different modality source images. Based on this model and our LZSC block, we propose an interpretable inverse fusion network named IFNet, which is utilized in the training of FNet.
\item FNet achieves leading results on eight datasets for five MMIF tasks: visible and infrared (VIS-IR), visible and near-infrared (VIS-NIR), computed tomography and magnetic resonance imaging (CT-MRI), positron emission tomography and MRI (PET-MRI), and single-photon emission computed tomography and MRI (SPECT-MRI) image fusion. 
Moreover, FNet also significantly enhances downstream object detection and \textcolor[rgb]{ 0,  0,  0}{and semantic segmentation} on VIS-IR image pairs.
\end{enumerate}

The remaining paper is structured as follows. In Section \ref{sec_2}, we review the sparse coding methods and prior works on the MMIF task and the inverse fusion process. Section \ref{sec_3} describes our proposed LZSC block, FNet,  IFNet, and the training process in detail. We conduct extensive experiments in Section \ref{sec_4} to validate our proposed method. Finally, Section \ref{sec_5} concludes the paper.

\section{Background}
\label{sec_2}
First, we review the sparse coding methods. Next, we discuss the prior works on the MMIF task. Finally, we review the inverse fusion process.
\begin{figure*} 
    \centering
    \includegraphics[width=0.9\linewidth]{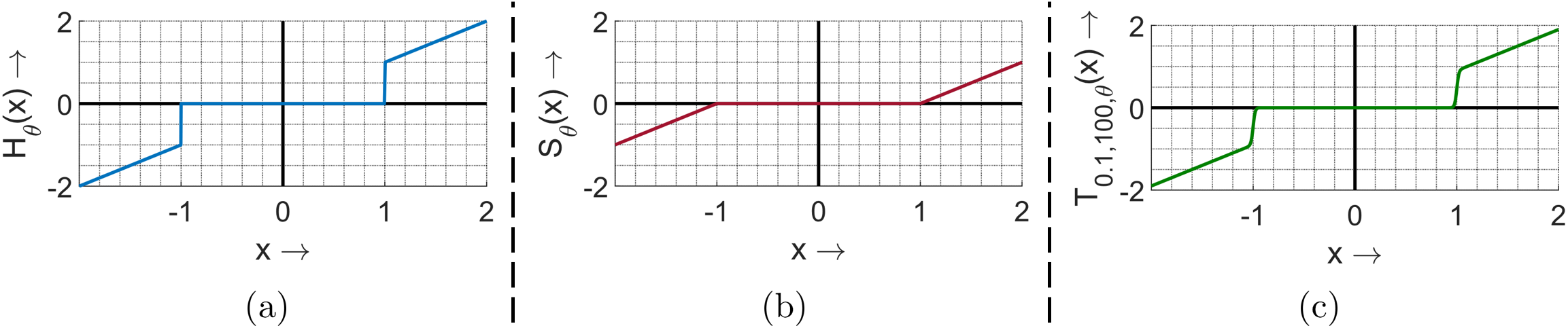}
  \caption{Different thresholding functions with threshold value $\theta$ as $1$. (a) Hard thresholding function, $H_{\theta}(\cdot)$, (b) Soft thresholding function, $S_{\theta}(\cdot)$, (c) Sigmoidal thresholding function, $T_{0.1,100,\theta}(\cdot)$.}
  \label{fig2} 
\end{figure*}

\subsection{Sparse Coding}
\label{sc}
Sparse coding (SC) is a widely used technique to select the salient features in a signal or image \cite{sparse1,sparse2,cunet}. The classical SC method represents a signal $x \in \mathbb{R}^{n \times 1}$ encoded into its sparse representation $z \in \mathbb{R}^{m \times 1}$ using a learned dictionary $D \in \mathbb{R}^{n \times m}$. The sparse representation $z$ is estimated by solving the $\ell_0$-regularized problem,
\begin{equation}\label{eq_z}
\underset{z}{\mathrm{Argmin}}\:\frac{1}{2}||\:x-Dz||_2^2 +\lambda ||z||_0 
\end{equation}

By imposing the $\ell_0$ regularization, $z$ is enforced to have only the salient features as the non-zero elements. However, Eqn. \ref{eq_z} is a non-convex and NP-hard formulation \cite{nphard}. A popular way to address this problem is to relax the non-convex $\ell_0$ pseudo-norm with the convex $\ell_1$ norm \cite{lcsc}. But only under certain conditions, the $\ell_1$-regularized sparse estimation matches with the $\ell_0$-regularized solution \cite{sparse}. Solving the original $\ell_0$-regularized problem can be more effective in many cases.
Blumensath \textit{et al.} \cite{ihta} proposed an iterative hard thresholding algorithm (IHTA) to solve Eqn. \ref{eq_z}, and \cite{normalized} modified IHTA to a normalized IHTA (NIHTA) for better convergence. Following NIHTA, the iteration step for updating $z$ is,

\begin{equation}\label{eq_ihta}
z^{k+1}=H_\theta\Big(z^k- \mu D^T\Big(Dz^k-x\Big)\Big)
\end{equation}

where $z^k$ is the sparse estimation at $k^{th}$ iteration and $\mu$ is the step size. $H_\theta(\cdot)$ is the hard thresholding function with threshold value $\theta$ defined as,

\begin{equation}\label{hard}
H_\theta(x) = 
\begin{cases}
  \:0 & \text{\:if\:\:}|x| \leq \theta \\
  \:x & \text{otherwise}
\end{cases}    
\end{equation}

Fig. \ref{fig2}-(a) shows the $H_\theta(\cdot)$ function, which promotes a sparse solution by nullifying the input with absolute values smaller than $\theta$. However, traditional iterative approaches take many iterations to solve the sparse estimation, and an algorithm unrolling-based neural network can have better estimation accuracy \cite{maximal}. Also, network formulation can facilitate task-specific optimization. Based on the IHTA algorithm,
\cite{encoder, maximal} modeled feed-forward neural networks for solving the classical SC problem. 

Generally, solving the classical SC problem is computationally intensive due to the matrix multiplication operation. Because of this, for high-resolution images, most methods split the image into overlapped patches, and each patch is processed independently to estimate the sparse representations and then aggregated using an averaging operation. Since the correlation between the patches is not considered, the estimation accuracy is poor. To address this issue, the convolutional sparse coding (CSC) method was proposed in \cite{csc_cvpr,csc_iccv}. CSC models the complete image as a sum over convolutional sparse representations (CSRs). For $\ell_0$-regularized CSC, the matrix multiplication in Eqn. \ref{eq_z} is substituted with the convolutional operation,
\begin{equation}\label{eq_csc}
\underset{z}{\mathrm{Argmin}}\:\frac{1}{2}||\:x-D(z)||_2^2 +\lambda ||z||_0 
\end{equation}

where $D(\cdot)$ is a convolution operation. Rodr{\'\i}guez \cite{csc} solved the $\ell_0$-regularized convolutional sparse coding (CSC) problem with an escape strategy-based iterative algorithm. However, there is no work that designed a neural network to solve the $\ell_0$-regularized CSC problem. 

The existing learning-based CSC method \cite{lcsc} solves the $\ell_1$-regularized CSC problem. Sreter \textit{et al.} \cite{lcsc} introduced a learned convolutional sparse coding (LCSC) block by unrolling a convolutional extension of ISTA \cite{ista}. The iteration steps of ISTA are similar to Eqn. \ref{eq_ihta}, only $H_\theta(\cdot)$ is replaced by the soft thresholding function, $S_\theta(\cdot)$ given by,

\begin{equation}\label{soft}
S_\theta(x)=max(|x|-\theta, 0)\:  sgn(x)   
\end{equation}

$sgn(\cdot)$ denotes the sign function. As illustrated in Fig. \ref{fig2}-(b), $S_\theta(\cdot)$ also nullifies the smaller input values like $H_\theta(\cdot)$. But, it also reduces the input with higher absolute values and thus over-penalizes them. This makes the magnitude of the sparse estimation lower than the $\ell_0$-regularized optimal sparse solution \cite{glista}, which is an inherent limitation of the ISTA-based methods. Due to this, the ISTA-based LCSC block \cite{lcsc} has limited performance in estimating the sparse solution.

To address the above-mentioned issues, we introduce a learnable $\ell_0$-regularized sparse coding (LZSC) block, which is developed by unrolling an iterative algorithm that solves the $\ell_0$-regularized CSC problem in Eqn. \ref{eq_csc}. The iterative algorithm is designed by incorporating Nesterov's momentum \cite{nesterov1983method} with a convolutional extension of the NIHTA steps in Eqn. \ref{eq_ihta}. However, the $H_\theta(\cdot)$ function in Eqn. \ref{eq_ihta} is a discontinuous function, and training neural networks with such a function is generally difficult \cite{encoder}. It is desirable to use smooth operator which can well approximate the non-smooth hard thresholding function \cite{algorithm}. To address this issue, we use a smooth sigmoidal thresholding function proposed in \cite{threshold}, 
\begin{equation}\label{sigmoidal}
T_{\alpha,\gamma,\theta}(x)=\frac{|x|-\alpha\theta}{1+e^{-\gamma(|x|-\theta)}}    
\end{equation}

where $\gamma$ is a parameter controlling the speed of the threshold transition, and $\alpha\in[0,1]$ indicates the additive adjustment that is made for input with larger absolute values. In the limiting cases, $T_{1,\infty,\theta}(\cdot)$ approximates $S_\theta(\cdot)$, and $T_{0,\infty,\theta}(\cdot)$ approximates $H_\theta(\cdot)$. Fig. \ref{fig2}-(c) shows $T_{0.1,100,\theta}(\cdot)$, which has a smoother transition compared to $H_\theta(\cdot)$ and unlike $S_\theta(\cdot)$, imposes negligible penalty for the larger input values. Instead of $H_\theta(\cdot)$, we use this $T_{0.1,100,\theta}(\cdot)$ for better network training. Our LZSC block-based method has superior performance than the existing LCSC block-based method \cite{lcsc} for the MMIF task.
\subsection{Prior Works on MMIF}
In this era of DL, the methods to solve the MMIF task can be categorized into two primary groups: pure DL-based models \cite{auifnet, U2fusion, SwinFusion, murf, laph, cddfuse, mda, itfuse, crossfuse,emma}, and algorithm unrolling-based models \cite{lrrnet, cunet}. Pure DL-based MMIF models typically utilize convolutional neural networks (CNNs) \cite{auifnet, U2fusion, murf,mda}, transformers \cite{SwinFusion, cddfuse, laph,itfuse,crossfuse, emma}, and generative adversarial networks (GANs) \cite{coconet} to construct a deep neural network to learn the mapping between the fused image and the source images. Although pure DL-based models have the potential to produce high-quality fused images, they empirically construct the the network using a trial and test strategy and often lack interpretability in the underlying fusion process. 

Whereas algorithm unrolling-based MMIF models \cite{cunet,lrrnet} derive inspiration from traditional algorithms and unroll an iterative algorithm into an interpretable deep neural network. LRRNet \cite{lrrnet} decomposes the source images into sparse and low-rank components based on a low-rank representation model and then combines the components to get the fused image. However, while decomposing the source images, it does not consider the dependency across modalities. Deng \textit{et al.} \cite{cunet} proposed a multi-modal convolutional sparse coding (MCSC) model to represent the MMIF process. Based on this model, they designed an interpretable network named CUNet that first separates the unique and common features from the different modality source images and then combines the features to get the fused image. In CUNet, the unique and common features are estimated using the LCSC block \cite{lcsc} that solves an $\ell_1$-regularized CSC problem. However, as discussed in Section \ref{sc}, the LCSC block-based method has limitations in estimating the sparse solution, and solving the $\ell_0$ regularized optimization problem can be highly effective here. In our work, we propose an $\ell_0$ regularized MCSC model to represent the relationship between different modalities, and based on this model and our LZSC block,  propose an interpretable MMIF network named FNet. Moreover, for the MMIF task, where the ground truth fused image is not available, considering an inverse fusion process can be very effective, which we discuss in the next subsection.
\subsection{Inverse Fusion Process}
\label{inverse}
In DL-based MMIF methods, one line of works \cite{sdnet, emma} considers an inverse fusion process, where the fused image is decomposed into the source images. As the quality of the decomposed source images is dependent on the quality of the fused image, constraining them to be similar to the original source images helps enforce the fused image to contain maximum information from the source images. However, for the inverse fusion task, \cite{sdnet,emma} used purely DL-based networks where interpretability of the underlying inverse fusion process is absent. In our work, we also consider an inverse fusion network named IFNet in the training of our proposed FNet to improve the fused image quality.  However, unlike other inverse fusion networks, our IFNet is derived from a novel $\ell_0$ regularized MCSC model. Such a model-based development gives interpretability of the inverse fusion process. Incorporating IFNet into the training of FNet significantly improves the fusion performance.

\section{Proposed Method}
\label{sec_3}
\subsection{Problem Statement}
In the MMIF process, different modality sensors capture the images of the same scene, which leads to the sharing of some common image features. However, since different modality sensors are used, the images also contain modality-specific unique features. The fused image can be obtained by combining these unique and common features. We consider two modalities of source images: $I_1 \in \mathbb{R}^{H \times W \times 1}$ and $I_2 \in \mathbb{R}^{H \times W \times 1}$, and fused image $I_f \in \mathbb{R}^{H \times W \times 1}$. $H$ and $W$ are the image height and width, respectively. The source images $I_1$ and $I_2$ share the same common feature $c\in \mathbb{R}^{H \times W \times K}$, where $K$ is the feature channel dimension. Moreover, $I_1$ and $I_2$ have their unique features  $u_1 \in \mathbb{R}^{H \times W \times K}$ and $u_2\in \mathbb{R}^{H \times W \times K}$ respectively. To select the salient features, we constrain $c$, $u_1$, and $u_2$ to be $\ell_0$-regularized. We propose the following  $\ell_0$-regularized MCSC model to represent the fusion process,

\begin{equation} \label{eq_fusion}
\begin{split}
&\mathrm{min}\:\: ||c||_0,\:\: ||u_1||_0,\:\: ||u_2||_0\\
s.t., \\
&I_1 = D_{c_1}(c)+D_{u_1}(u_1) \\
&I_2 = D_{c_2}(c)+D_{u_2}(u_2)\\
&I_f = \underbrace{G_c(c)}_{\substack{\text{common part} \\ \text{from } I_1 \text{ and } I_2}}+\underbrace{G_{u_1}(u_1)}_{\substack{\text{unique part} \\ \text{from } I_1}}+\underbrace{G_{u_2}(u_2)}_{\substack{\text{unique part} \\ \text{from } I_2}}
\end{split}
\end{equation}

where $D_{c_1}(\cdot)$, $D_{u_1}(\cdot)$, $D_{c_2}(\cdot)$, $D_{u_2}(\cdot)$, $G_c(\cdot)$, $G_{u_1}(\cdot)$ and $G_{u_2}(\cdot)$ denote learnable convolution operations.
Now, given the source images $I_1$ and $I_2$, we solve the following optimization problem to estimate $c$, $u_1$, and $u_2$,

\begin{equation} \label{eq_fus_opt}
\begin{split}
\underset{c,\: u_1,\: u_2}{\mathrm{Argmin}}\:&\frac{1}{2}\Big|\Big|\:I_1-D_{c_1}(c)-D_{u_1}(u_1)\Big|\Big|_2^2\\
+&\frac{1}{2}\Big|\Big|\:I_2-D_{c_2}(c)-D_{u_2}(u_2)\Big|\Big|_2^2\\+& \lambda_c ||c||_0+\lambda_{u_1}||u_1||_0+\lambda_{u_2}||u_2||_0
\end{split}
\end{equation}

Following \cite{cunet}, we first update $u_1$ and $u_2$ by setting the value of $c$ to zero. Then, $c$ is updated by fixing $u_1$ and $u_2$. The steps are:

\begin{itemize}
\item $u_i, i\in\{1,2\}$ is updated by solving the equation,
\begin{equation}\label{eq_u1}
\underset{u_i}{\mathrm{Argmin}}\:\frac{1}{2}\Big|\Big|\:I_i-D_{u_i}(u_i)\Big|\Big|_2^2 +\lambda_{u_i} ||u_i||_0 
\end{equation}
\item $c$ is updated by solving the equation,
\begin{equation}\label{eq_c1}
\begin{split}
\underset{c}{\mathrm{Argmin}}\:&\frac{1}{2}\Big|\Big|\:\hat{I}_1-D_{c_1}(c)\Big|\Big|_2^2 \\+&\frac{1}{2}\Big|\Big|\:\hat{I}_2-D_{c_2}(c)\Big|\Big|_2^2 +\lambda_c ||c||_0 
\end{split}
\end{equation}

where $\hat{I}_1 = I_1-D_{u_1}(u_1)$ and $\hat{I}_2 = I_2-D_{u_2}(u_2)$. By channel-wise concatenating $\hat{I}_1$ and $\hat{I}_2$ to $\hat{I}_{1,2}\in \mathbb{R}^{H \times W \times 2}$, Eqn. \ref{eq_c1} can be written as,

\begin{equation}\label{eq_c}
\underset{c}{\mathrm{Argmin}}\:\frac{1}{2}\Big|\Big|\:\hat{I}_{1,2}-L_c(c)\Big|\Big|_2^2 +\lambda ||c||_0 
\end{equation}

here $L_c(\cdot)$ is learnable convolution operation. 
\end{itemize}

As discussed in Section \ref{inverse}, employing an inverse fusion process in the training of the fusion network can enhance the fused image quality. Motivated by this, we consider an inverse fusion process where the fused image is separated into the source images. Since the fused image has information content from the different modality source images, we can represent the fused image as a combination of features from the source images. We consider that the fused image $I_f$ has the features $x_1 \in \mathbb{R}^{H \times W \times K}$ and $x_2\in \mathbb{R}^{H \times W \times K}$ corresponding to the source images $I_1$ and $I_2$ respectively. To get the salient features, we constrain $x_1$ and $x_2$ to be $\ell_0$-regularized. We propose the following $\ell_0$-regularized MCSC model to represent the inverse fusion process,

\begin{equation} \label{eq_invfusion}
\begin{split}
&\mathrm{min}\:\: ||x_1||_0,\:\: ||x_2||_0\\
s.t., \\
&I_f = \underbrace{G_{x_1}(x_1)}_{\substack{\text{from } I_1}}+\underbrace{G_{x_2}(x_2)}_{\substack{\text{from } I_2}}\\
&I_1 = D_{x_1}(x_1)\\
&I_2 = D_{x_2}(x_2)
\end{split}
\end{equation}

where $G_{x_1}(\cdot)$, $G_{x_2}(\cdot)$, $D_{x_1}(\cdot)$ and $D_{x_2}(\cdot)$ denote learnable convolution operations. 
Given the fused image $I_f$, we estimate $x_1$ and $x_2$ by solving the follwing optimization problem,

\begin{equation} \label{eq_invfus_opt}
\begin{split}
\underset{x_1,\: x_2}{\mathrm{Argmin}}\:&\frac{1}{2}\Big|\Big|\:I_f-G_{x_1}(x_1)-G_{x_2}(x_2)\Big|\Big|_2^2\\+& \lambda_{x_1}||x_1||_0+\lambda_{x_2}||x_2||_0
\end{split}
\end{equation}

For solving Eqn. \ref{eq_invfus_opt}, we update $x_1$ by setting $x_2$ to zero, and vice versa. The updation equation for $x_i, i \in \{1,2\}$ is,
\begin{equation}\label{eq_x1}
\underset{x_i}{\mathrm{Argmin}}\:\frac{1}{2}\Big|\Big|\:I_f-G_{x_i}(x_i)\Big|\Big|_2^2 +\lambda_{x_i} ||x_i||_0 
\end{equation}

Eqns. \ref{eq_u1}, \ref{eq_c} and \ref{eq_x1} are $\ell_0$ regularized CSC problems, and to solve them we introduce a novel LZSC block described below.
\begin{figure*} 
    \centering
  \includegraphics[width=0.68\linewidth]{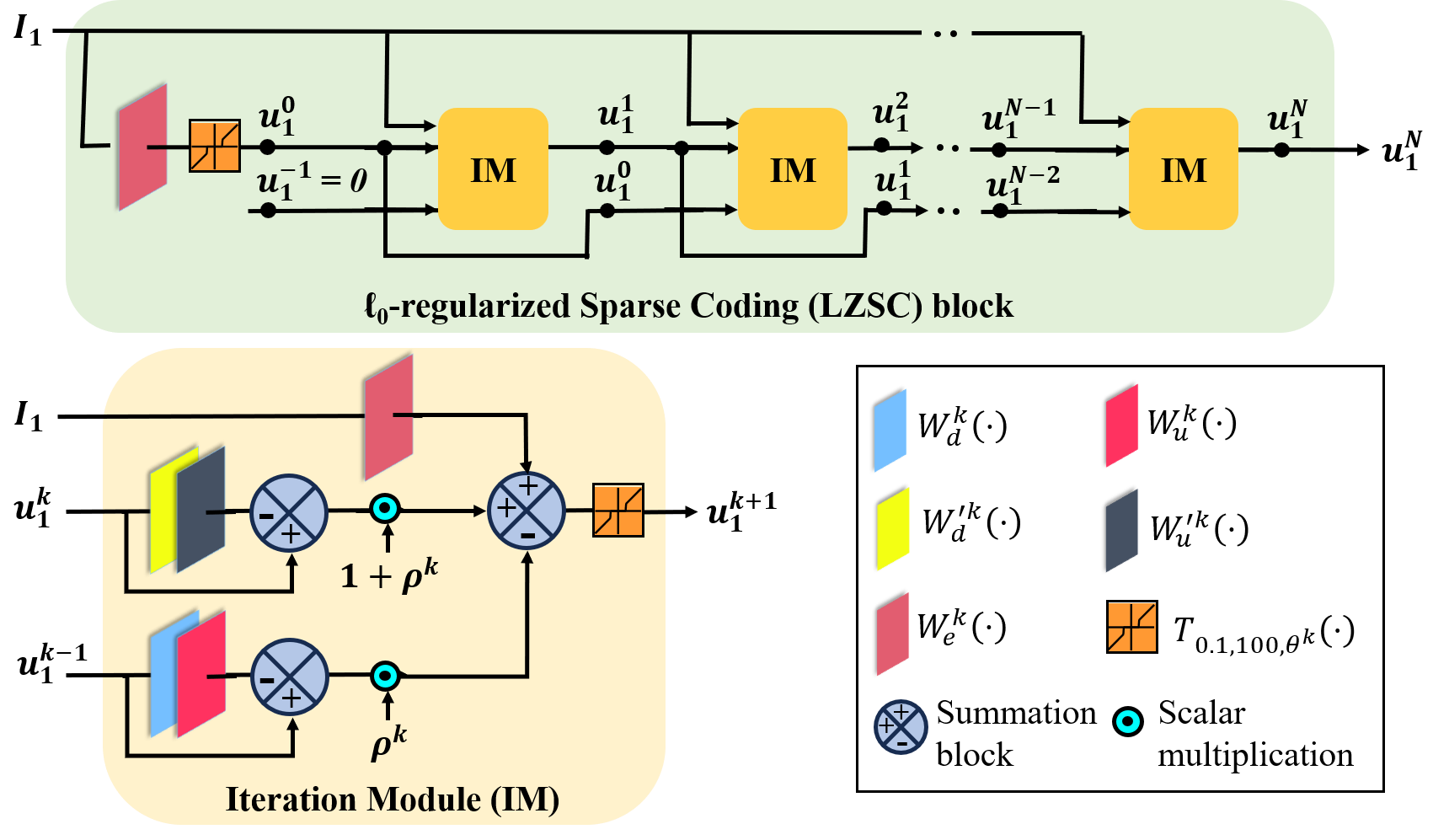}
  
  \caption{Architecture of proposed LZSC block and the structure of iteration module (IM). Given the input image $I_1$, the LZSC block estimates the $\ell_0$-regularized sparse feature $u_1^N$ at $N^{th}$ iteration. The structure of the LZSC block for estimating the sparse features $c$, $u_2$, $x_1$, and $x_2$ is the same; only the input image is different.}
  
  \label{fig3} 
\end{figure*}
\subsection{Solving the $\ell_0$-regularized CSC problem}
\label{fiht}
We consider the $\ell_0$ regularized CSC problem for the updation of $u_1$ (Eqn. \ref{eq_u1}),
\begin{equation}\nonumber
\underset{u_1}{\mathrm{Argmin}}\:\frac{1}{2}\Big|\Big|\:I_1-D_u(u_1)\Big|\Big|_2^2 +\lambda_{u_1} ||u_1||_0 
\end{equation}

To solve this, inspired by \cite{lcsc}, we first propose a convolutional extension of the NIHTA iteration step in Eqn \ref{eq_ihta}. Thus, the iteration step for updating $u_1$ becomes,

\begin{equation}\label{eq_ihta1}
u_1^{k+1}=H_\theta\Big(u_1^k- W_u\Big(W_d\Big(u_1^k\Big)-I_1\Big)\Big)
\end{equation}

where $u_1^k$ is the sparse estimation at $k^{th}$ iteration, $W_u(\cdot)$ and $W_d(\cdot)$ are learnable convolution layers, and $\theta$ is a learnable parameter. However, in Eqn. \ref{eq_ihta1}, the current estimation $u_1^{k+1}$ only depends on the previous estimation $u_1^{k}$. To improve the convergence, we can introduce Neterov's momentum \cite{nesterov1983method} which considers the previous two estimation steps $u_1^{k}$ and $u_1^{k-1}$ to update the current estimation. Inspired by \cite{fista}, we introduce Nesterov's momentum in Equation \ref{eq_ihta1} to update $u_1$ as,
\begin{equation} \label{eq_fihta1}
\begin{split}
u_1^{k}&=H_\theta\Big(v^k- W_u\Big(W_d\Big(v^k\Big)\Big)+W_u\Big(I_1\Big)\Big)\\
v^{k+1}&=u_1^{k}+\rho^k\big(u_1^{k}-u_1^{k-1}\big)
\end{split}
\end{equation}

where $\rho^k$, a learnable parameter, is the update weight at each iteration and should increase monotonously with iteration number $k$. Now combining the two steps in Eqn. \ref{eq_fihta1}, we get,
\begin{equation} \label{eq_lfihta2}
\begin{split}
u_1^{k+1}=H_{\theta}\Big(&\Big(1+\rho^k\Big)\Big(u_1^k-W_u\Big(W_d\Big(u_1^k\Big)\Big)\Big)\\&
- \rho^k\Big(u_1^{k-1}-W_u\Big(W_d\Big(u_1^{k-1}\Big)\Big)\Big)\\&+W_u\Big(I_1\Big)\Big)
\end{split}
\end{equation}

In Eqn. \ref{eq_lfihta2}, the convolutional layers are the same for all the iterations. However, having different learnable layers and parameters at each iteration usually improves the performance \cite{maximal}. Moreover, the estimation accuracy improves if we have different convolutional layers for the input $I_1$ \cite{adalista}. Motivated by these, we update the iteration step as,

\begin{equation} \label{eq_lfihta}
\begin{split}
u_1^{k+1}=H_{\theta^{k}}\Big(&\Big(1+\rho^k\Big)\Big(u_1^k-W_u^k\Big(W_d^k\Big(u_1^k\Big)\Big)\Big)\\
&- \rho^k\Big(u_1^{k-1}-W_u^{k}\Big(W_d^{k}\Big(u_1^{k-1}\Big)\Big)\Big)\\&+W_e^k\Big(I_1\Big)\Big)
\end{split}
\end{equation}

$W_u^k(\cdot)$, $W_d^k(\cdot)$ and $W_e^k(\cdot)$ are learnable convolution layers at each iteration. Moreover, we propose to use different convolution layers for $u_1^{k}$ and $u_1^{k-1}$ for better learning,

\begin{equation} \label{eq_lfihta3}
\begin{split}
u_1^{k+1}=H_{\theta^{k}}\Big(&\Big(1+\rho^k\Big)\Big(u_1^k-W_u^k\Big(W_d^k\Big(u_1^k\Big)\Big)\Big)\\
&- \rho^k\Big(u_1^{k-1}-W_u^{'k}\Big(W_d^{'k}\Big(u_1^{k-1}\Big)\Big)\Big)\\&+W_e^k\Big(I_1\Big)\Big)
\end{split}
\end{equation}

$W_u^{'k}(\cdot)$ and $W_d^{'k}(\cdot)$ are learnable convolution layers at each iteration. As discussed in Section \ref{sc}, we replace the discontinuous thresholding function $H_{\theta^{k}}(\cdot)$ with a continuous function $T_{0.1,100,\theta^{k}}(\cdot)$ for better network training. Finally, the iteration step for updating $u_1$ becomes,

\begin{equation} \label{eq_lfihtaf}
\begin{split}
u_1^{k+1}=T_{0.1,100,\theta^{k}}\Big(&\Big(1+\rho^k\Big)\Big(u_1^k-W_u^k\Big(W_d^k\Big(u_1^k\Big)\Big)\Big)\\
&- \rho^k\Big(u_1^{k-1}-W_u^{'k}\Big(W_d^{'k}\Big(u_1^{k-1}\Big)\Big)\Big)\\&+W_e^k\Big(I_1\Big)\Big)
\end{split}
\end{equation}

In Eqn. \ref{eq_lfihtaf}, $\theta^k$ and $\rho^k$ are learnable, and they may learn non-positive values, which contradicts their definition. Moreover, $\rho^k$ should be within $0$ to $1$ and increase with the iteration number $k$ for better convergence. Also, $\theta^k$ should decrease with the iteration number $k$ as the sparse estimation accuracy improves with $k$. Inspired by \cite{fistanet}, we constrain $\theta^k$ and $\rho^k$ as,
\begin{equation} \label{eq_constrain}
\begin{split}
&\theta^{k}=sp(w_\theta k+b_\theta)\:\:,\:\: w_\theta <0\\
&\rho^k=\frac{sp(w_\rho k+b_\rho)-sp(b_\rho)}{sp(w_\rho k+b_\rho)}\:\:,\:\:  w_\rho <0
\end{split}
\end{equation}

$sp(\cdot)$ denotes the softplus function. Here, $w_\theta$, $b_\theta$, $w_\rho$ and $b_\rho$ are the parameters which are learned. 

 By unfolding the iteration steps in Eqn. \ref{eq_lfihtaf}, we design the iteration module (IM) shown in Fig. \ref{fig3}. 
Moreover, we assume $u_1^k=0; \:\: \forall k<0$, and stack multiple IMs to construct the LZSC block. 
We utilize this LZSC block to design our fusion network FNet and inverse fusion network IFNet, the details of which are described in the next subsection.

\begin{figure} 
    \centering
  \includegraphics[width=0.85\linewidth]{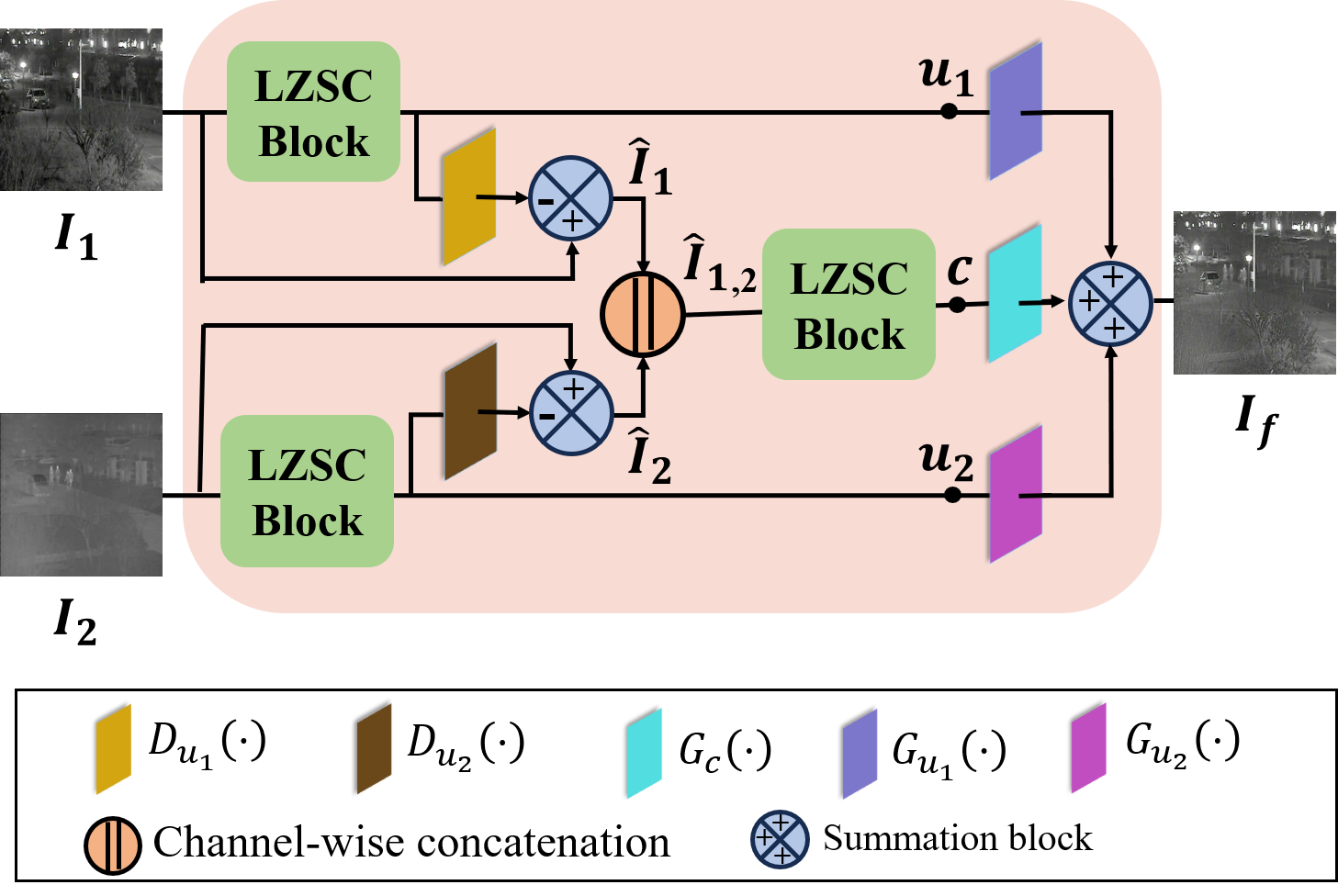}
  
  \caption{Architecture of proposed FNet. Given the source images $I_1$ and $I_2$, FNet generates the fused image $I_f$.}
  
  \label{fig4} 
\end{figure}
\begin{figure} 
    \centering
  \includegraphics[width=0.6\linewidth]{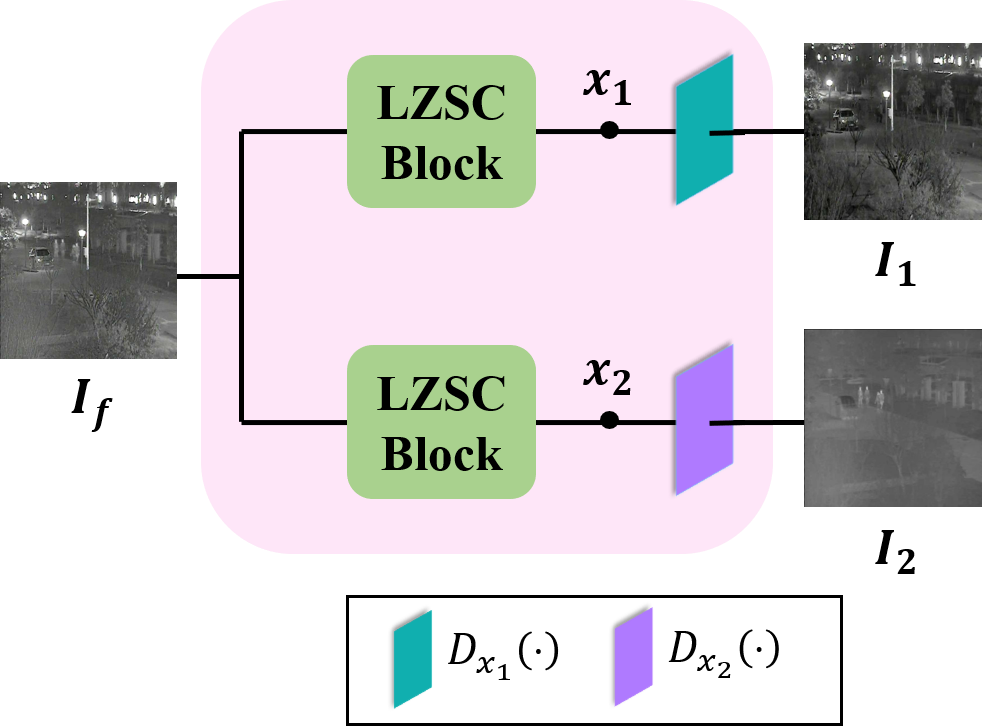}
  
  \caption{Architecture of proposed IFNet. Given the fused image $I_f$, IFNet generates the source images $I_1$ and $I_2$.}
  
  \label{fig5} 
\end{figure}
\subsection{Network Architectures}
\label{FNet}
The proposed FNet architecture is illustrated in Fig. \ref{fig4}. Given the source images $I_1$ and $I_2$ from different modalities, first the unique features $u_1$ and $u_2$ are estimated using two LZSC blocks. Then we get $\hat{I}_1$ and $\hat{I}_2$ as,

\begin{equation}
\hat{I}_1 = I_1-D_{u_1}(u_1) ,  \:\: \hat{I}_2 = I_2-D_{u_2}(u_2)  
\end{equation}

Following this, $\hat{I}_1$ and $\hat{I}_2$ are channel-wise concatenated to get $\hat{I}_{1,2}$. Then, the common feature $c$ is estimated from $\hat{I}_{1,2}$ using one LZSC block. Finally, we generate the fused image $I_f$ from the unique and common features as,

\begin{equation}
I_f = G_c(c)+G_{u_1}(u_1)+G_{u_2}(u_2)   
\end{equation}

The proposed IFNet architecture is also illustrated in Fig. \ref{fig5}. Given the fused image $I_f$, first, we estimate $x_1$ and $x_2$, the features corresponding to $I_1$ and $I_2$ using two LZSC blocks. Then, $I_1$ and $I_2$ are obtained by,
\begin{equation}
I_1 = D_{x_1}(x_1) ,  \:\:  
I_2 = D_{x_2}(x_2)
\end{equation}

\begin{table*} [h!]
	\fontsize{7.5}{8}\selectfont
	\centering
	\caption{Performance comparison for VIS-IR, VIS-NIR, CT-MRI, PET-MRI, and SPECT-MRI image fusion tasks. FLOPs and runtime values are reported under the setting of fusing two source images of resolution $320\times 320$. We highlight the best and second-best performances in \textcolor{red}{\textbf{red}} and \textcolor{blue}{\textbf{blue}} colors, respectively. $\uparrow$ means high value, and $\downarrow$ means low value desired.}
	\begin{tabular}{c|r|APFADDABEDBA|c}
		\toprule
		\multicolumn{1}{c|}{\multirow{7}[6]{*}{\makecell{Tasks \\ (Datasets)}}} & \multirow{2}[2]{*}{Models} & \text{AUIFNet} & \text{SwinFusion} & \text{U2Fusion} & \text{CoCoNet} & \text{LapH}  & \text{MURF}  & \text{LRRNet} & \text{CDDFuse} & \text{MDA}   &\text{ITFuse} & \text{CrossFuse} & \text{EMMA}  & \multicolumn{1}{c}{\multirow{4}[4]{*}{\makecell{FNet \\ (Ours)}}} \\
		&       & \text{\cite{auifnet}}  & \text{\cite{SwinFusion}}   & \text{\cite{U2fusion}}  & \text{\cite{coconet}}   & \text{\cite{laph}}  & \text{\cite{murf}}  & \text{\cite{lrrnet}}  & \text{\cite{cddfuse}}  & \text{\cite{mda}}  & \text{\cite{itfuse}}  & \text{\cite{crossfuse}}  & \text{\cite{emma}}  &  \\
		\cmidrule{2-14}          & Publications & \text{TCSVT} & \text{IEEE}  & \text{TPAMI} & \text{IJCV}  & \text{TCSVT} & \text{TPAMI} & \text{TPAMI} & \text{CVPR}  & \text{NC}    & \text{PR}    & \text{IF}    & \text{CVPR}  &  \\
		& Years & \text{2021}  & \text{2022}  & \text{2022}  & \text{2023}  & \text{2023}  & \text{2023}  & \text{2023}  & \text{2023}  & \text{2024}  & \text{2024}  & \text{2024}  & \text{2024}  &  \\
		\cmidrule{2-15}          & Params (K) & {12}    & \text{974}   & \text{659}   & \text{6845}  & \text{134}   & \text{116}   & \text{49}    & \text{1188}  & \text{516}   & \text{82}    & \text{23154} & \text{1518}  & \text{420} \\
		& FLOPs (G) & {4.83}  & \text{117.53} & \text{135.06} & \text{66.79} & \text{3.16}  & \text{184.36} & \text{5.19}  & \text{184.56} & \text{51.66} & \text{8.42}  & \text{55.86} & \text{13.89} & \text{0.33} \\
		& Runtime (s) & \text{0.023} & \text{0.728} & \text{0.019} & \text{0.010} & \text{0.028} & \text{0.155} & \text{0.051} & \text{0.096} & \text{0.017} & \text{0.032} & \text{0.925} & \text{0.041} & \text{0.051} \\
		\toprule
		\multicolumn{1}{c|}{\multirow{6}[2]{*}{\makecell{VIS-IR \\ (TNO)}}} & MI ↑  & \text{1.58}  & \textcolor[rgb]{ 0,  0,  1}{\textbf{2.26}} & \text{1.37}  & \text{1.54}  & \text{1.30}  & \text{1.40}  & \text{1.76}  & \text{2.19}  & \text{1.40}  & \text{1.52}  & \text{2.18}  & \text{2.12}  & \textcolor[rgb]{ 1,  0,  0}{\textbf{2.57}} \\
		& \textcolor[rgb]{ 0,  0,  0}{CE ↓}    & \textcolor[rgb]{ 0,  0,  1}{\textbf{2.20}} & \textcolor[rgb]{ 1,  0,  0}{\textbf{2.14}} & \text{3.04}  & \text{2.50}  & \text{2.63}  & \text{2.63}  & \text{2.99}  & \text{3.29}  & \text{5.92}  & \text{2.23}  & \text{3.23}  & \text{3.08}  & \textcolor[rgb]{ 0,  0,  1}{\textbf{2.20}} \\
		& VIF ↑ & \text{0.61}  & \text{0.75}  & \text{0.58}  & \text{0.64}  & \text{0.63}  & \text{0.50}  & \text{0.54}  & \textcolor[rgb]{ 0,  0,  1}{\textbf{0.77}} & \text{0.37}  & \text{0.45}  & \text{0.73}  & \text{0.70}  & \textcolor[rgb]{ 1,  0,  0}{\textbf{0.80}} \\
		& Qabf ↑ & \text{0.43}  & \text{0.53}  & \text{0.44}  & \text{0.32}  & \text{0.44}  & \text{0.38}  & \text{0.37}  & \textcolor[rgb]{ 0,  0,  1}{\textbf{0.54}} & \text{0.15}  & \text{0.21}  & \text{0.46}  & \text{0.49}  & \textcolor[rgb]{ 1,  0,  0}{\textbf{0.57}} \\
		& \textcolor[rgb]{ 0,  0,  0}{Qcb ↑}   & \text{0.48}  & \text{0.50}  & \textcolor[rgb]{ 0,  0,  1}{\textbf{0.51}} & \text{0.43}  & \text{0.46}  & \text{0.48}  & \text{0.48}  & \textcolor[rgb]{ 0,  0,  1}{\textbf{0.51}} & \text{0.30}  & \text{0.45}  & \text{0.48}  & \textcolor[rgb]{ 0,  0,  1}{\textbf{0.51}} & \textcolor[rgb]{ 1,  0,  0}{\textbf{0.52}} \\
		& SSIM ↑ & \text{0.93}  & \textcolor[rgb]{ 0,  0,  1}{\textbf{1.04}} & \text{0.99}  & \text{0.72}  & \text{0.85}  & \text{0.98}  & \text{0.84}  & \text{1.03}  & \text{0.48}  & \text{0.78}  & \text{0.90}  & \text{0.97}  & \textcolor[rgb]{ 1,  0,  0}{\textbf{1.05}} \\
		\midrule
		\multicolumn{1}{c|}{\multirow{6}[2]{*}{\makecell{VIS-IR \\ (Road- \\ Scene)}}} & MI ↑  & \text{1.96}  & \textcolor[rgb]{ 0,  0,  1}{\textbf{2.34}} & \text{1.87}  & \text{1.83}  & \text{1.88}  & \text{1.86}  & \text{1.99}  & \text{2.30}  & \text{1.73}  & \text{1.80}  & \text{2.32}  & \text{2.27}  & \textcolor[rgb]{ 1,  0,  0}{\textbf{2.56}} \\
		& \textcolor[rgb]{ 0,  0,  0}{CE ↓}    & \text{2.27}  & \text{1.92}  & \text{2.07}  & \text{2.34}  & \text{2.24}  & \text{2.28}  & \text{3.74}  & \text{1.91}  & \text{3.09}  & \textcolor[rgb]{ 1,  0,  0}{\textbf{1.32}} & \text{3.27}  & \text{2.21}  & \textcolor[rgb]{ 0,  0,  1}{\textbf{1.81}} \\
		& VIF ↑ & \text{0.64}  & \text{0.67}  & \text{0.60}  & \text{0.57}  & \text{0.66}  & \text{0.55}  & \text{0.49}  & \textcolor[rgb]{ 0,  0,  1}{\textbf{0.69}} & \text{0.44}  & \text{0.43}  & \text{0.60}  & \text{0.66}  & \textcolor[rgb]{ 1,  0,  0}{\textbf{0.70}} \\
		& Qabf ↑ & \textcolor[rgb]{ 0,  0,  1}{\textbf{0.51}} & \text{0.49}  & \textcolor[rgb]{ 0,  0,  1}{\textbf{0.51}} & \text{0.37}  & \text{0.50}  & \text{0.47}  & \text{0.35}  & \textcolor[rgb]{ 1,  0,  0}{\textbf{0.52}} & \text{0.20}  & \text{0.21}  & \text{0.37}  & \text{0.47}  & \textcolor[rgb]{ 0,  0,  1}{\textbf{0.51}} \\
		& \textcolor[rgb]{ 0,  0,  0}{Qcb ↑}   & \text{0.47}  & \textcolor[rgb]{ 0,  0,  1}{\textbf{0.49}} & \textcolor[rgb]{ 1,  0,  0}{\textbf{0.50}} & \text{0.48}  & \textcolor[rgb]{ 1,  0,  0}{\textbf{0.50}} & \text{0.47}  & \textcolor[rgb]{ 1,  0,  0}{\textbf{0.50}} & \textcolor[rgb]{ 0,  0,  1}{\textbf{0.49}} & \text{0.34}  & 0.46  & \textcolor[rgb]{ 0,  0,  1}{\textbf{0.49}} & \textcolor[rgb]{ 1,  0,  0}{\textbf{0.50}} & \text{0.48} \\
		& SSIM ↑ & \text{0.96}  & \textcolor[rgb]{ 0,  0,  1}{\textbf{0.99}} & \text{0.97}  & \text{0.74}  & \text{0.87}  & \text{0.96}  & \text{0.64}  & \text{0.98}  & \text{0.65}  & \text{0.68}  & \text{0.73}  & \text{0.91}  & \textcolor[rgb]{ 1,  0,  0}{\textbf{1.01}} \\
		\midrule
		\multicolumn{1}{c|}{\multirow{6}[2]{*}{\textcolor[rgb]{ 0,  0,  0}{\makecell{VIS-IR \\ (CATS)}}}} & \textcolor[rgb]{ 0,  0,  0}{MI ↑}   & \text{2.16}  & \text{2.72}  & \text{1.78}  & \text{1.69}  & \text{1.74}  & \text{1.57}  & \text{2.22}  & \textcolor[rgb]{ 1,  0,  0}{\textbf{2.77}} & \text{1.70}  & \text{2.06}  & \text{2.35}  & \text{2.53}  & \textcolor[rgb]{ 0,  0,  1}{\textbf{2.75}} \\
		& \textcolor[rgb]{ 0,  0,  0}{CE ↓}    & \text{3.82}  & \text{2.81}  & \text{3.09}  & \text{3.26}  & \text{3.42}  & \text{3.90}  & \text{4.18}  & \text{4.54}  & \text{7.32}  & \textcolor[rgb]{ 0,  0,  1}{\textbf{2.78}} & \text{3.48}  & \text{4.34}  & \textcolor[rgb]{ 1,  0,  0}{\textbf{2.75}} \\
		& \textcolor[rgb]{ 0,  0,  0}{VIF ↑}  & \text{0.70}  & \textcolor[rgb]{ 0,  0,  1}{\textbf{0.87}} & \text{0.62}  & \text{0.53}  & \text{0.80}  & \text{0.47}  & \text{0.63}  & \textcolor[rgb]{ 1,  0,  0}{\textbf{0.89}} & \text{0.38}  & \text{0.56}  & \text{0.78}  & \text{0.81}  & \textcolor[rgb]{ 1,  0,  0}{\textbf{0.89}} \\
		& \textcolor[rgb]{ 0,  0,  0}{Qabf ↑}  & \text{0.54}  & \textcolor[rgb]{ 0,  0,  1}{\textbf{0.66}} & \text{0.51}  & \text{0.39}  & \text{0.62}  & \text{0.30}  & \text{0.55}  & \text{0.65}  & \text{0.15}  & \text{0.24}  & \text{0.61}  & \text{0.65}  & \textcolor[rgb]{ 1,  0,  0}{\textbf{0.68}} \\
		& \textcolor[rgb]{ 0,  0,  0}{Qcb ↑}   & \text{0.50}  & \textcolor[rgb]{ 0,  0,  1}{\textbf{0.54}} & \text{0.53}  & \text{0.49}  & \text{0.52}  & \text{0.41}  & \text{0.52}  & \text{0.52}  & \text{0.28}  & \text{0.45}  & \textcolor[rgb]{ 0,  0,  1}{\textbf{0.54}} & \textcolor[rgb]{ 1,  0,  0}{\textbf{0.56}} & \textcolor[rgb]{ 0,  0,  1}{\textbf{0.54}} \\
		& \textcolor[rgb]{ 0,  0,  0}{SSIM ↑}  & \text{0.78}  & \textcolor[rgb]{ 1,  0,  0}{\textbf{0.95}} & \text{0.88}  & \text{0.66}  & \text{0.83}  & \text{0.79}  & \text{0.72}  & \textcolor[rgb]{ 1,  0,  0}{\textbf{0.95}} & \text{0.39}  & \text{0.71}  & \text{0.85}  & \textcolor[rgb]{ 0,  0,  1}{\textbf{0.90}} & \textcolor[rgb]{ 1,  0,  0}{\textbf{0.95}} \\
		\midrule
		\multicolumn{1}{c|}{\multirow{6}[2]{*}{\textcolor[rgb]{ 0,  0,  0}{\makecell{VIS-IR \\ (KAIST)}}}} & \textcolor[rgb]{ 0,  0,  0}{MI ↑}  & \text{1.98}  & \textcolor[rgb]{ 1,  0,  0}{\textbf{3.01}} & \text{1.79}  & \text{1.72}  & \text{1.71}  & \text{1.81}  & \text{1.93}  & \text{2.98}  & \text{1.76}  & \text{2.06}  & \text{2.51}  & \text{2.87}  & \textcolor[rgb]{ 0,  0,  1}{\textbf{2.99}} \\
		& \textcolor[rgb]{ 0,  0,  0}{CE ↓}   & \text{2.30}  & \text{2.29}  & \textcolor[rgb]{ 1,  0,  0}{\textbf{1.37}} & \textcolor[rgb]{ 0,  0,  1}{\textbf{1.78}} & \text{2.17}  & \text{2.23}  & \text{3.48}  & \text{2.45}  & \text{11.03} & \text{2.04}  & \text{4.27}  & \text{2.78}  & \text{2.63} \\
		& \textcolor[rgb]{ 0,  0,  0}{VIF ↑} & \text{0.60}  & \textcolor[rgb]{ 0,  0,  1}{\textbf{0.84}} & \text{0.62}  & \text{0.64}  & \text{0.73}  & \text{0.47}  & \text{0.62}  & \textcolor[rgb]{ 1,  0,  0}{\textbf{0.86}} & \text{0.43}  & \text{0.55}  & \text{0.74}  & \text{0.82}  & \textcolor[rgb]{ 0,  0,  1}{\textbf{0.84}} \\
		& \textcolor[rgb]{ 0,  0,  0}{Qabf ↑} & \text{0.42}  & \textcolor[rgb]{ 0,  0,  1}{\textbf{0.62}} & \text{0.47}  & \text{0.36}  & \text{0.53}  & \text{0.30}  & \text{0.46}  & \text{0.61}  & \text{0.19}  & \text{0.27}  & \text{0.57}  & \text{0.60}  & \textcolor[rgb]{ 1,  0,  0}{\textbf{0.63}} \\
		& \textcolor[rgb]{ 0,  0,  0}{Qcb ↑}  & \textcolor[rgb]{ 1,  0,  0}{\textbf{0.51}} & \textcolor[rgb]{ 1,  0,  0}{\textbf{0.51}} & \textcolor[rgb]{ 1,  0,  0}{\textbf{0.51}} & \text{0.45}  & \text{0.45}  & \text{0.42}  & \text{0.46}  & \text{0.48}  & \text{0.21}  & \text{0.40}  & \text{0.41}  & \textcolor[rgb]{ 1,  0,  0}{\textbf{0.51}} & \textcolor[rgb]{ 0,  0,  1}{\textbf{0.50}} \\
		& \textcolor[rgb]{ 0,  0,  0}{SSIM ↑} & \text{0.70}  & \textcolor[rgb]{ 0,  0,  1}{\textbf{0.96}} & \text{0.91}  & \text{0.67}  & \text{0.80}  & \text{0.82}  & \text{0.82}  & \textcolor[rgb]{ 1,  0,  0}{\textbf{0.97}} & \text{0.41}  & \text{0.78}  & \text{0.86}  & \text{0.94}  & \textcolor[rgb]{ 1,  0,  0}{\textbf{0.97}} \\
		\midrule
		\multicolumn{1}{c|}{\multirow{6}[2]{*}{\makecell{VIS-NIR \\ (RGB-NIR \\ Scene)}}} & MI ↑  & \textcolor[rgb]{ 0,  0,  1}{\textbf{2.23}} & \text{2.14}  & \text{1.91}  & \text{1.90}  & \text{1.87}  & \text{1.73}  & \text{2.11}  & \textcolor[rgb]{ 1,  0,  0}{\textbf{2.28}} & \text{1.69}  & \text{2.14}  & \text{2.21}  & \text{2.20}  & \text{2.09} \\
		& \textcolor[rgb]{ 0,  0,  0}{CE ↓}    & \text{2.08}  & \text{1.91}  & \text{2.28}  & \text{2.03}  & \text{3.02}  & \text{2.67}  & \text{3.03}  & \textcolor[rgb]{ 1,  0,  0}{\textbf{1.10}} & \text{3.92}  & \text{2.35}  & \text{2.50}  & \text{2.28}  & \textcolor[rgb]{ 0,  0,  1}{\textbf{1.87}} \\
		& VIF ↑ & \text{0.61}  & \text{0.62}  & \text{0.53}  & \text{0.52}  & \text{0.60}  & \text{0.45}  & \text{0.49}  & \textcolor[rgb]{ 1,  0,  0}{\textbf{0.64}} & \text{0.23}  & \text{0.46}  & \textcolor[rgb]{ 1,  0,  0}{\textbf{0.64}}  & \text{0.60}  & \textcolor[rgb]{ 0,  0,  1}{\textbf{0.63}} \\
		& Qabf ↑ & \text{0.38}  & \text{0.42}  & \text{0.40}  & \text{0.26}  & \text{0.38}  & \text{0.33}  & \text{0.35}  & \textcolor[rgb]{ 0,  0,  1}{\textbf{0.44}} & \text{0.05}  & \text{0.20}  & \textcolor[rgb]{ 1,  0,  0}{\textbf{0.45}}  & \text{0.42}  & \textcolor[rgb]{ 1,  0,  0}{\textbf{0.45}} \\
		& \textcolor[rgb]{ 0,  0,  0}{Qcb ↑}   & \text{0.49}  & \text{0.50}  & \textcolor[rgb]{ 1,  0,  0}{\textbf{0.52}} & \text{0.49}  & \text{0.48}  & \text{0.50}  & \textcolor[rgb]{ 1,  0,  0}{\textbf{0.52}} & \textcolor[rgb]{ 0,  0,  1}{\textbf{0.51}} & \text{0.29}  & \text{0.46}  & \text{0.50}  & \textcolor[rgb]{ 1,  0,  0}{\textbf{0.52}} & \textcolor[rgb]{ 0,  0,  1}{\textbf{0.51}} \\
		& SSIM ↑ & \text{1.11}  & \text{1.16}  & \text{1.19}  & \text{0.83}  & \text{0.95}  & \text{1.00}  & \text{0.97}  & \textcolor[rgb]{ 1,  0,  0}{\textbf{1.21}} & \text{0.36}  & \text{0.98}  & \text{1.14}  & \text{1.13}  & \textcolor[rgb]{ 0,  0,  1}{\textbf{1.20}} \\
		\midrule
		\multicolumn{1}{c|}{\multirow{6}[2]{*}{\makecell{CT-MRI \\ (Harvard \\ medical)}}} & MI ↑  & \textcolor[rgb]{ 1,  0,  0}{\textbf{2.36}} & \text{2.26}  & \text{2.08}  & \text{2.15}  & \text{1.99}  & \text{2.10}  & \text{2.12}  & \text{2.26}  & \text{2.09}  & \text{2.15}  & \textcolor[rgb]{ 0,  0,  1}{\textbf{2.30}}  & \text{2.27} & \text{2.23} \\
		& \textcolor[rgb]{ 0,  0,  0}{CE ↓}    & \text{6.38}  & \textcolor[rgb]{ 0,  0,  1}{\textbf{0.22}} & \text{0.87}  & \text{1.54}  & \text{1.49}  & \text{0.79}  & \text{16.52} & \text{1.30}  & \text{17.55} & \textcolor[rgb]{ 1,  0,  0}{\textbf{0.20}} & \text{15.66} & \text{1.32}  & \textcolor[rgb]{ 1,  0,  0}{\textbf{0.20}} \\
		& VIF ↑ & \text{0.41}  & \textcolor[rgb]{ 0,  0,  1}{\textbf{0.56}} & \text{0.37}  & \text{0.39}  & \text{0.44}  & \text{0.38}  & \text{0.38}  & \text{0.50}  & \text{0.36}  & \text{0.39}  & \text{0.48}  & \text{0.49}  & \textcolor[rgb]{ 1,  0,  0}{\textbf{0.57}} \\
		& Qabf ↑ & \text{0.48}  & \text{0.58}  & \text{0.46}  & \text{0.44}  & \text{0.52}  & \text{0.36}  & \text{0.34}  & \textcolor[rgb]{ 1,  0,  0}{\textbf{0.60}} & \text{0.14}  & \text{0.16}  & \text{0.37}  & \text{0.55}  & \textcolor[rgb]{ 0,  0,  1}{\textbf{0.59}} \\
		& \textcolor[rgb]{ 0,  0,  0}{Qcb ↑}   & \text{0.57}  & \textcolor[rgb]{ 1,  0,  0}{\textbf{0.68}} & \text{0.32}  & \text{0.20}  & \text{0.50}  & \text{0.26}  & \text{0.23}  & \textcolor[rgb]{ 0,  0,  1}{\textbf{0.64}} & \text{0.13}  & \text{0.31}  & \text{0.20}  & \text{0.56}  & \textcolor[rgb]{ 1,  0,  0}{\textbf{0.68}} \\
		& SSIM ↑ & \text{0.46}  & \textcolor[rgb]{ 1,  0,  0}{\textbf{1.34}} & \text{0.49}  & \text{0.44}  & \text{0.86}  & \text{0.48}  & \text{0.37}  & \textcolor[rgb]{ 1,  0,  0}{\textbf{1.34}} & \text{0.31}  & \text{0.39}  & \text{0.39}  & \textcolor[rgb]{ 0,  0,  1}{\textbf{0.97}} & \textcolor[rgb]{ 1,  0,  0}{\textbf{1.34}} \\
		\midrule
		\multicolumn{1}{c|}{\multirow{6}[2]{*}{\makecell{PET-MRI \\ (Harvard \\ medical)}}} & MI ↑  & \text{1.84}  & \textcolor[rgb]{ 0,  0,  1}{\textbf{1.96}} & \text{1.69}  & \text{1.60}  & \text{1.81}  & \text{1.57}  & \text{1.56}  & \text{1.88}  & \text{1.61}  & \text{1.65}  & \text{1.82}  & \text{1.75}  & \textcolor[rgb]{ 1,  0,  0}{\textbf{1.97}} \\
		& \textcolor[rgb]{ 0,  0,  0}{CE ↓}    & \text{0.63}  & \text{0.78}  & \text{0.61}  & \text{1.64}  & \text{3.26}  & \textcolor[rgb]{ 0,  0,  1}{\textbf{0.22}} & \text{21.03} & \text{0.91}  & \text{21.91} & \textcolor[rgb]{ 1,  0,  0}{\textbf{0.15}} & \text{18.78} & \text{1.16}  & \text{0.78} \\
		& VIF ↑ & \text{0.37}  & \textcolor[rgb]{ 0,  0,  1}{\textbf{0.70}} & \text{0.40}  & \text{0.41}  & \text{0.52}  & \text{0.45}  & \text{0.37}  & \text{0.63}  & \text{0.48}  & \text{0.49}  & \text{0.59}  & \text{0.59}  & \textcolor[rgb]{ 1,  0,  0}{\textbf{0.71}} \\
		& Qabf ↑ & \text{0.31}  & \text{0.64}  & \text{0.49}  & \text{0.47}  & \text{0.57}  & \text{0.16}  & \text{0.21}  & \textcolor[rgb]{ 0,  0,  1}{\textbf{0.65}} & \text{0.16}  & \text{0.20}  & \text{0.34}  & \text{0.57}  & \textcolor[rgb]{ 1,  0,  0}{\textbf{0.66}} \\
		& \textcolor[rgb]{ 0,  0,  0}{Qcb ↑}   & \text{0.60}  & \textcolor[rgb]{ 1,  0,  0}{\textbf{0.71}} & \text{0.27}  & \text{0.22}  & \text{0.44}  & \text{0.21}  & \text{0.17}  & \textcolor[rgb]{ 0,  0,  1}{\textbf{0.68}} & \text{0.11}  & \text{0.28}  & \text{0.22}  & \text{0.57}  & \textcolor[rgb]{ 1,  0,  0}{\textbf{0.71}} \\
		& SSIM ↑ & \text{0.29}  & \textcolor[rgb]{ 0,  0,  1}{\textbf{1.48}} & \text{1.39}  & \text{0.34}  & \text{0.77}  & \text{0.33}  & \text{0.22}  & \text{1.47}  & \text{0.27}  & \text{0.35}  & \text{0.36}  & \text{0.97}  & \textcolor[rgb]{ 1,  0,  0}{\textbf{1.50}} \\
		\midrule
		\multicolumn{1}{c|}{\multirow{6}[2]{*}{\makecell{SPECT- \\ MRI \\ (Harvard \\ medical)}}} & MI ↑  & \text{1.81}  & \textcolor[rgb]{ 1,  0,  0}{\textbf{1.94}} & \text{1.68}  & \text{1.67}  & \text{1.76}  & \text{1.73}  & \text{1.63}  & \text{1.86}  & \text{1.67}  & \text{1.75}  & \text{1.80}  & \text{1.83}  & \textcolor[rgb]{ 0,  0,  1}{\textbf{1.91}} \\
		& \textcolor[rgb]{ 0,  0,  0}{CE ↓}    & \text{1.20}  & \text{1.13}  & \text{0.96}  & \text{2.00}  & \text{4.18}  & \textcolor[rgb]{ 0,  0,  1}{\textbf{0.84}} & \text{20.89} & \text{1.25}  & \text{22.24} & \textcolor[rgb]{ 1,  0,  0}{\textbf{0.83}} & \text{16.30} & \text{1.44}  & \text{1.11} \\
		& VIF ↑ & \text{0.39}  & \textcolor[rgb]{ 0,  0,  1}{\textbf{0.61}} & \text{0.48}  & \text{0.44}  & \text{0.57}  & \text{0.50}  & \text{0.34}  & \text{0.60}  & \text{0.44}  & \text{0.47}  & \text{0.58}  & \text{0.55}  & \textcolor[rgb]{ 1,  0,  0}{\textbf{0.63}} \\
		& Qabf ↑ & \text{0.32}  & \text{0.62}  & \text{0.57}  & \text{0.44}  & \text{0.60}  & \text{0.24}  & \text{0.20}  & \textcolor[rgb]{ 0,  0,  1}{\textbf{0.65}} & \text{0.12}  & \text{0.18}  & \text{0.29}  & \text{0.58}  & \textcolor[rgb]{ 1,  0,  0}{\textbf{0.66}} \\
		& \textcolor[rgb]{ 0,  0,  0}{Qcb ↑}   & \text{0.61}  & \textcolor[rgb]{ 0,  0,  1}{\textbf{0.68}} & \text{0.24}  & \text{0.18}  & \text{0.41}  & \text{0.20}  & \text{0.17}  & \text{0.66}  & \text{0.08}  & \text{0.24}  & \text{0.18}  & \text{0.49}  & \textcolor[rgb]{ 1,  0,  0}{\textbf{0.69}} \\
		& SSIM ↑ & \text{0.27}  & \textcolor[rgb]{ 1,  0,  0}{\textbf{1.49}} & \text{1.41}  & \text{0.29}  & \text{0.75}  & \text{0.32}  & \text{0.21}  & \textcolor[rgb]{ 0,  0,  1}{\textbf{1.48}} & \text{0.19}  & \text{0.32}  & \text{0.33}  & \text{0.66}  & \textcolor[rgb]{ 1,  0,  0}{\textbf{1.49}} \\
		\bottomrule
	\end{tabular}%
	\label{tab:vif}
\end{table*}
\subsection{Training Process}
\label{training}
\begin{figure} 
    \centering
  \includegraphics[width=0.9\linewidth]{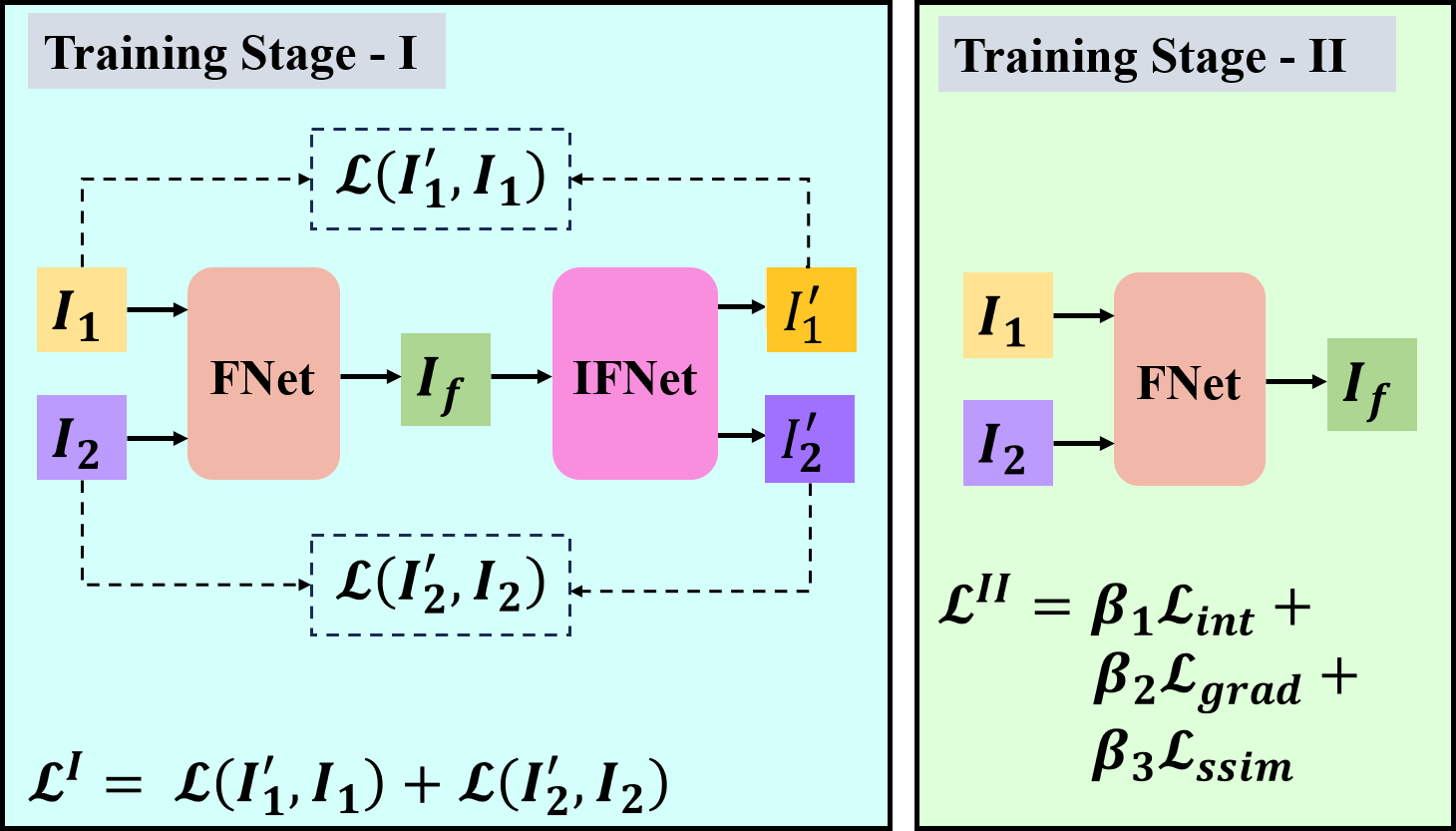}
  
  \caption{Pipeline for two-stage training of our proposed FNet.}
  \label{fig10} 
\end{figure}

In the MMIF task, where the ground truth fused image is not available, a two-stage training procedure can be very effective \cite{cddfuse}. Motivated by this, we propose a two-stage training procedure for FNet. Fig. \ref{fig10} shows the two-stage training pipeline. The details are described below.

\noindent
\textbf{Training stage I.} We consider both the fusion and inverse fusion processes in the training stage I. The idea is that the original source images should be the same as the source images generated in the inverse fusion process. Given the source images $I_1$ and $I_2$, we generate the fused image $I_f$ using FNet. Then, $I_f$ is input to IFNet to generate the source images $I_1^{'}$ and $I_2^{'}$. The training constraints $I_1^{'}$ to be similar to $I_1$ and $I_2^{'}$  to be similar to $I_2$, using the loss function,

\begin{equation}\label{training1}
\mathcal{L}^I = \mathcal{L}(I_1^{'},I_1)+\mathcal{L}(I_2^{'},I_2)
\end{equation}

where, $\mathcal{L}(I_1^{'},I_1)=||\:I_1^{'}-I_1\:||_1 +  ||\:\nabla I_1^{'}-\nabla I_1\:||_1$. $\nabla$ is the Sobel gradient operator. Both FNet and IFNet are trained in the training stage I.

\noindent
\textbf{Training stage II.} We consider only the fusion process in the training stage II. Our motivation is to constrain the generated fused image to have maximum similarity with the source images. Here, the source images $I_1$ and $I_2$ are fed to a nearly well-trained FNet to generate $I_f$. Inspired by \cite{SwinFusion}, the total loss is.
\begin{equation}\label{training2}
\mathcal{L}^{II} = \beta_1  \mathcal{L}_{int} +\beta_2 \mathcal{L}_{grad} +\beta_3 \mathcal{L}_{ssim}
\end{equation}

where $\mathcal{L}_{int}=||\:I_f-max(I_1,I_2)\:||_1$, $\mathcal{L}_{grad}=||\:\nabla I_f-max(\nabla I_1,\nabla I_2)\:||_1$. $\mathcal{L}_{ssim}=w_1 (1-ssim(I_1,I_f))+w_2 (1-ssim(I_2,I_f))$. $ssim(\cdot)$ is the structural similarity index measure between two images. $w_1=\frac{\nabla I_1}{\nabla I_1+\nabla I_2}$, $w_2=\frac{\nabla I_2}{\nabla I_1+\nabla I_2}$. $\beta_1$, $\beta_2$ and $\beta_3$ are tuning parameters. 

\section{Experiments}
\label{sec_4}
\begin{figure*} 
    \centering
  \includegraphics[width=0.88\linewidth]{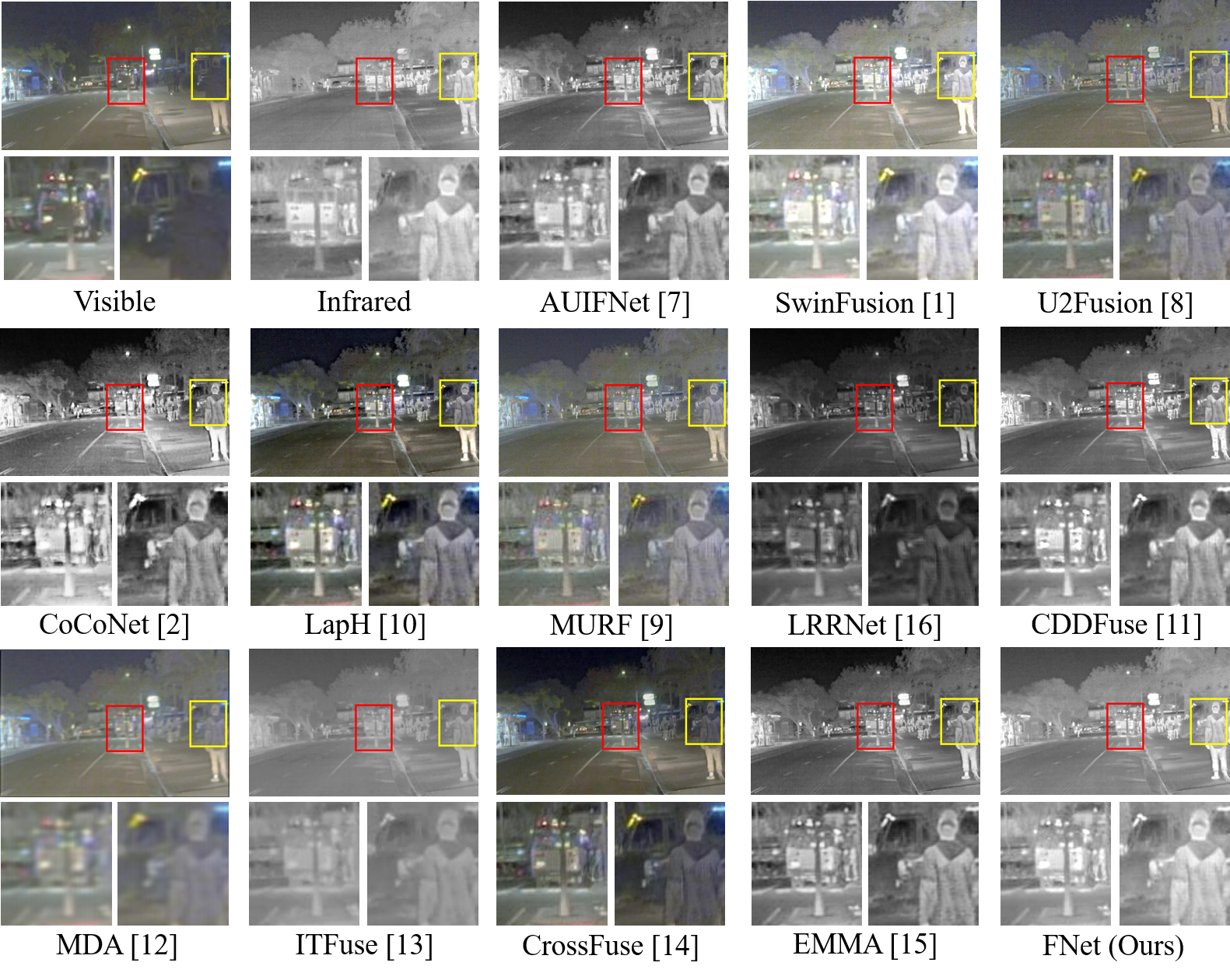}
  
  \caption{Visual comparison with the SOTA methods for the VIS-IR image fusion task on the RoadScene dataset. Compared to SwinFusion \cite{SwinFusion} and CDDFuse \cite{cddfuse}, our proposed FNet better preserves the structure information of the source images.}
  \label{fig6} 
\end{figure*}
\begin{figure*} 
    \centering
  \includegraphics[width=0.88\linewidth]{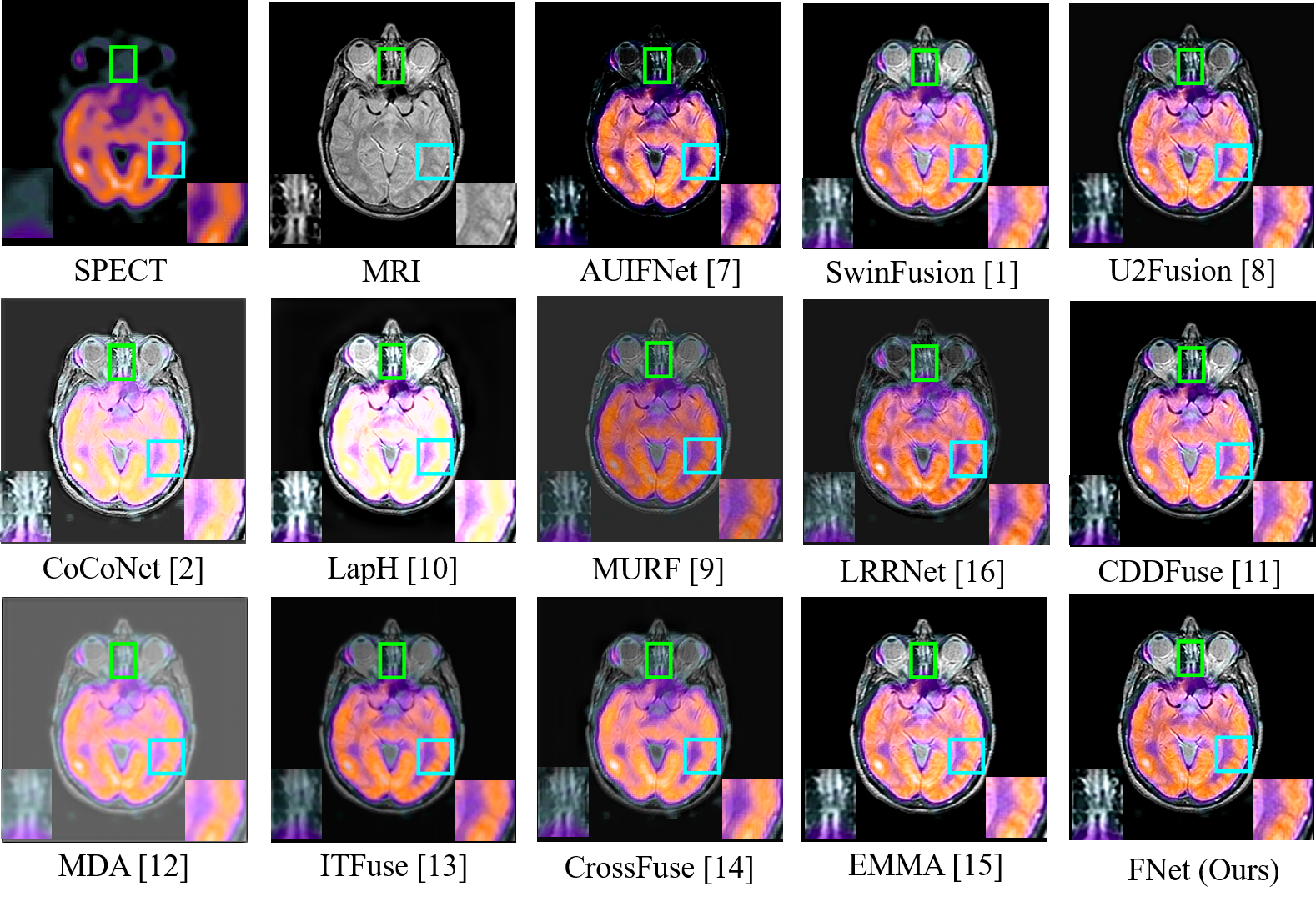}
  
  \caption{Visual comparison with the SOTA methods for the MRI-SPECT image fusion task on Harvard medical dataset. Compared to SwinFusion \cite{SwinFusion} and CDDFuse \cite{cddfuse}, our FNet better preserves the structure information of the source images.}
  \label{fig7} 
\end{figure*}
The performance of our proposed FNet is evaluated on five MMIF tasks: \textbf{i)} VIS-IR, \textbf{ii)} VIS-NIR, \textbf{iii)} CT-MRI, \textbf{iv)} PET-MRI, and \textbf{v)} SPECT-MRI image fusion. Section \ref{setup} presents the experimental setup, including implementation details, datasets, training settings, and evaluation metrics. Then, we report the quantitative and qualitative comparison results with the SOTA methods in Section \ref{sota}. The intermediate features are visualized in Section \ref{visualization} to show the good network interpretability of our FNet. \textcolor[rgb]{ 0,  0,  0}{Moreover, we compare the estimated features of FNet with the SOTA methods in Section \ref{interpretability}. Then, we perform experimental analysis in Section \ref{analyze_l0} to show that $\ell _0$-regularization is implemented in the three LZSC blocks of FNet.} In Section \ref{downstream}, we compare FNet with SOTA methods on downstream object detection \textcolor[rgb]{ 0,  0,  0}{and semantic segmentation} in VIS-IR image pairs. Finally, in Section \ref{ablation}, ablation experiments are conducted to demonstrate the effectiveness of our proposed method.  

\subsection{Experimental Setup}
\label{setup}

\noindent
\textbf{Implementation details.} In FNet, each convolution layer is set to have a kernel size $9\times 9$, and number of filters $K=64$. In the LZSC block, the number of IMs is set to $4$. We set the tuning parameters $\beta_1=20,\beta_2=20,\beta_3=15$ for the loss function in Eqn. \ref{training2}. 

\noindent
\textbf{Datasets.} For both training and testing, publicly available datasets are used. We train FNet with $1,444$ image pairs from the MSRS VIS-IR dataset \cite{SwinFusion} and test on other datasets without any fine-tuning to check the model's generalization ability. We test: \textbf{i)} VIS-IR task on $25$ image pairs from the TNO dataset \cite{tno}, $50$ image pairs from the RoadScene dataset \cite{U2fusion}, \textcolor[rgb]{ 0,  0,  0}{$65$ image pairs from the CATS dataset \cite{cats}, and $44$ image pairs from the KAIST dataset \cite{kaist}},  \textbf{ii)} VIS-NIR task on $20$ image pairs from the RGB-NIR Scene dataset \cite{nir}, \textbf{iii)} CT-MRI task on $21$ image pairs from the Harvard medical dataset \cite{harvard}, \textbf{iv)} PET-MRI task on $42$ image pairs from the Harvard medical dataset and \textbf{v)} SPECT-MRI task on $73$ image pairs from the Harvard medical dataset.

\noindent
\textbf{Training settings.} We train FNet in two stages. In both stages, the training is conducted with Adam optimizer with a constant learning rate of $1\times10^{-4}$ for $20,000$ iterations with a batch size of $16$. In each iteration, we randomly crop the image pairs to a patch size of $128\times 128$ and augment them by horizontal and vertical flipping. All experiments are conducted using an NVIDIA A40 GPU within the PyTorch framework. 

\noindent
\textbf{Evaluation metrics.} We objectively compare fusion performance with six metrics: mutual information (MI), \textcolor[rgb]{ 0,  0,  0}{cross-entropy (CE)}, visual information fidelity (VIF) \cite{vif}, edge information (Qabf) \cite{qabf}, \textcolor[rgb]{ 0,  0,  0}{quantified content-based metric (Qcb) \cite{qcb}}, and structural similarity index measure (SSIM) \cite{ssim}. A higher value of MI, VIF, Qabf, \textcolor[rgb]{ 0,  0,  0}{Qcb}, and SSIM \textcolor[rgb]{ 0,  0,  0}{and a lower value of CE} indicate superior fusion performance. We follow the calculations given in \cite{cddfuse,mmif_survey}. 
\subsection{Performance Comparison with the SOTA Methods}
\label{sota}
FNet is compared  with twelve recent SOTA methods: AUIFNet \cite{auifnet}, SwinFusion \cite{SwinFusion}, U2Fusion \cite{U2fusion}, CoCoNet \cite{coconet}, LapH \cite{laph}, MURF \cite{murf}, LRRNet \cite{lrrnet}, CDDFuse \cite{cddfuse}, MDA \cite{mda}, ITFuse \cite{itfuse}, CrossFuse \cite{crossfuse}, and EMMA \cite{emma}. For the SOTA methods, we use their trained network parameters and official code to generate the fused images.

\subsubsection{Quantitative Comparison}
Table \ref{tab:vif} shows the quantitative comparison for MMIF tasks on eight datasets. Along with the MI, CE, VIF, Qabf, Qcb, and SSIM metrics, we also list the model parameters, FLOPs, and average GPU runtime. FLOPs and average runtime values are reported under the setting of fusing two source images of resolution $320\times 320$. FNet has the leading performance on almost all six metrics across the eight datasets. The results demonstrate that FNet can preserve the essential structural information of the source images. Among the other SOTA methods, SwinFusion and CDDFuse have comparable performance with our method. It is worth noting that FNet has much lower parameters, FLOPs, and runtime than these two methods. \textcolor[rgb]{ 0,  0,  0}{Results in Table \ref{tab:vif} show the generalization ability of the MMIF methods. We have also conducted experiments to compare the performance of the methods when they are trained and tested on the same dataset. We have presented the experiments in Section V of the supplementary material.}

\subsubsection{Qualitative Comparison}
Fig. \ref{fig6} shows a visual comparison of FNet with the SOTA methods for the VIS-IR image fusion task on the RoadScene dataset. FNet successfully separates the common background in the VIS-IR image pairs, unique scene details in the visible image, and unique objects in the infrared image and combines all this information to generate the fused image. Compared to the SOTA methods, FNet can better preserve the structure and texture details of the source images in the fused image.

Fig. \ref{fig7} shows a visual comparison of the SPECT-MRI image fusion task on the Harvard medical dataset. 
Our FNet effectively separates the common edge details in the SPECT-MRI image pairs, the unique functional details in the SPECT image, and the tissue details in the MRI image and combines all these features into the fused image. Compared to the SOTA methods, FNet can better preserve the tissue and structure details of the source images.
More visual comparison results are given in Section III of the supplementary material.
\begin{figure*}[t!] 
    \centering
  \includegraphics[width=0.8\linewidth]{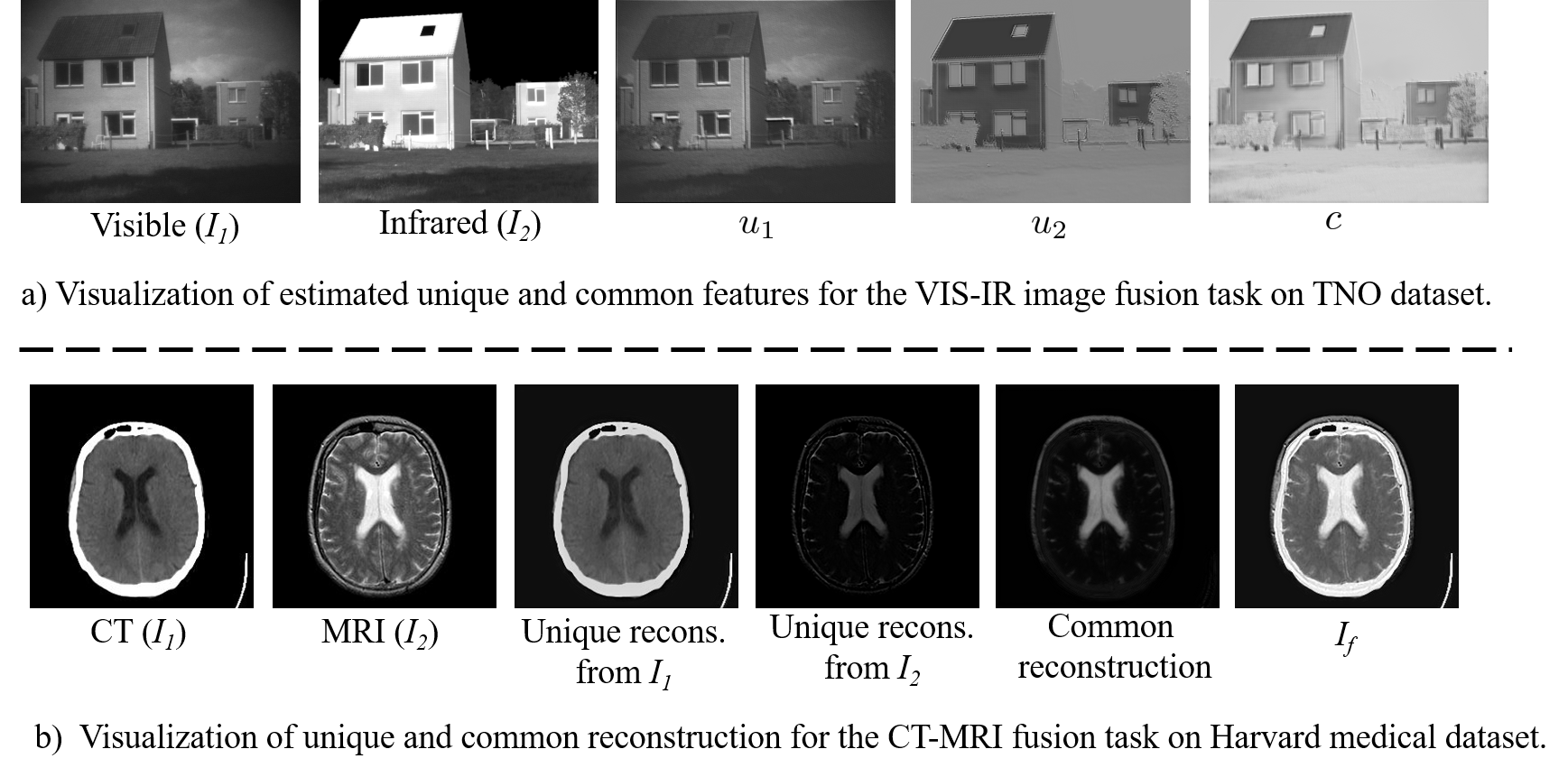}
  
  \caption{Visualization of intermediate results.}
  \label{fig8} 
\end{figure*}
\begin{figure*} 
    \centering
\includegraphics[width=0.9\linewidth]{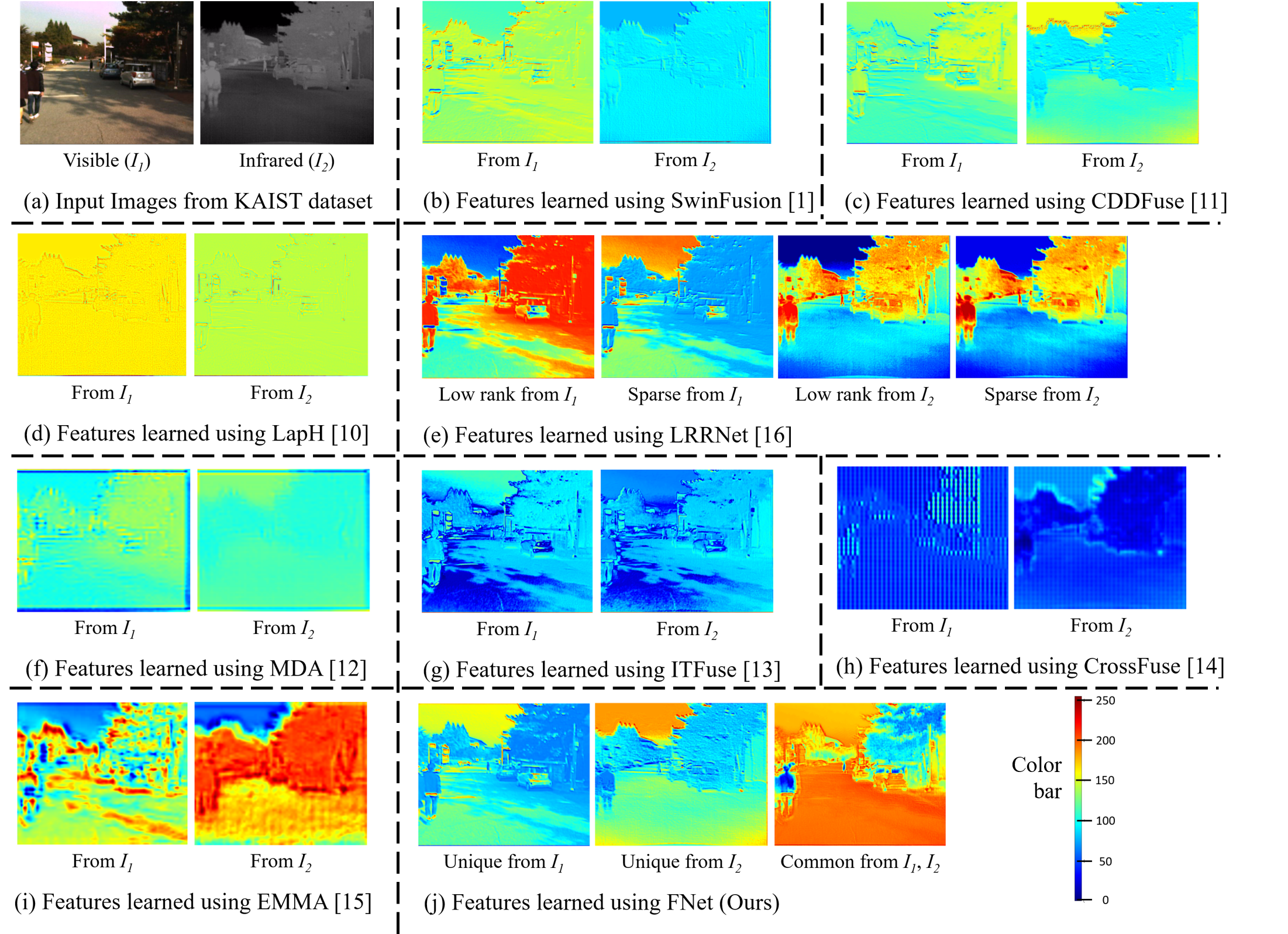}
  
  \caption{\textcolor[rgb]{ 0,  0,  0}{Visual comparison of intermediate features with SOTA methods.}}
  \label{fig_interprtability} 
\end{figure*}
\subsection{Visualization of Intermediate Results}
\label{visualization}
\subsubsection{Visualization of unique and common features}
In our MMIF network FNet, we use three LZSC blocks to estimate the unique and common features from the source images of different modalities. This provides interpretability of the underlying fusion process. Fig. \ref{fig8}-(a) shows the estimated unique and common features for the VIS-IR image fusion task on the TNO dataset. The unique extraction $u_1$ from the visible image ($I_1$) consists of the scene-specific details — such as sky texture. This information is not present in the infrared image ($I_2$). The unique extraction $u_2$ from the infrared image consists of the details of the thermal radiation, which does not exist in the visible image. Thus, the unique features of visible and infrared images are specific to their own modality. The common feature $c$ extracted from the two images is the edges of the different objects.

\subsubsection{Visualization of unique and common reconstruction}
In FNet, we first estimate the unique and common features. Then these features undergo convolution operations to obtain the unique and common reconstruction parts, which are then added to get the fused image. Fig. \ref{fig8}-(b) shows the unique and common reconstructions of the CT-MRI image fusion task on the Harvard medical dataset. As we can see, the unique reconstruction from the CT image ($I_1$) preserves the anatomical structures such as bone contours, whereas the unique part of the MRI image ($I_2$) has the soft tissue details. The common reconstruction retains only the overlapping
shapes shared by both scans; for instance, the thick skull structure visible in the CT image and the fine
tissue patterns from the MRI image are not included in this common reconstruction. The fused image $I_f$ consists of both the unique and common reconstruction parts. More visualization of intermediate results are given in Section IV of the supplementary material.

\begin{figure*}[t!]  
    \centering
  \includegraphics[width=0.95\linewidth]{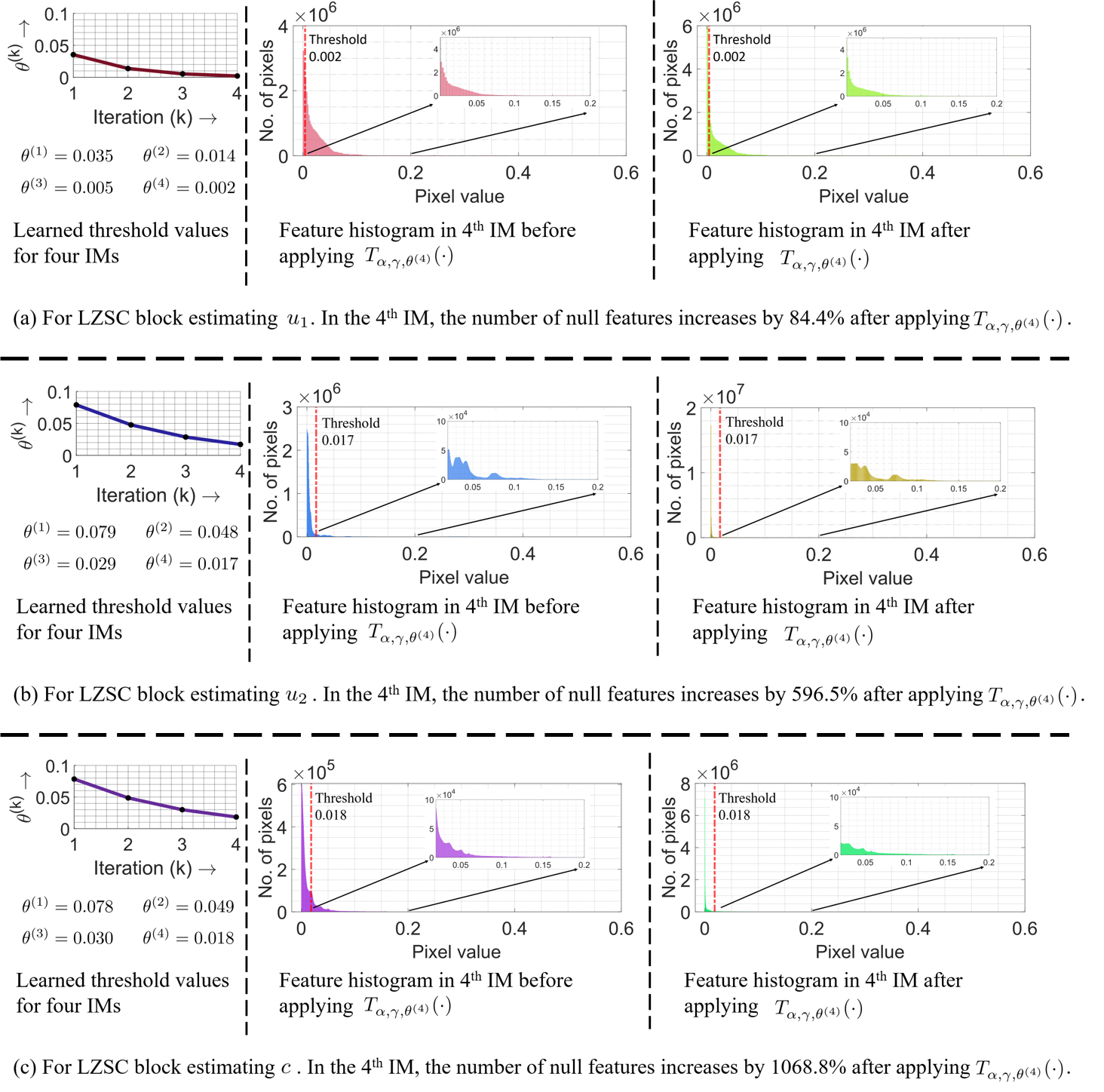}
  
  \caption{\textcolor[rgb]{ 0,  0,  0}{Comparing the histograms of absolute values of the features before and after applying   $T_{\alpha,\gamma,\theta}(\cdot)$. The features are for an input image pair from the TNO dataset. $T_{\alpha,\gamma,\theta}(\cdot)$ nullifies the features with absolute values lower than threshold, and imposes very small penalty for large features.}}
  \label{fig_threshold} 
\end{figure*}
\textcolor[rgb]{ 0,  0,  0}{\subsection{Comparison of Features with SOTA Methods}
\label{interpretability}
In this subsection, we compare the features estimated by FNet with the SOTA methods to explore the interpretability of the extracted features. Fig. \ref{fig_interprtability} shows the features estimated by different MMIF methods for a VIS-IR image pair from the KAIST dataset. For the SOTA methods, we display the features used to reconstruct the fused image. We compare our method with pure DL-based approaches: SwinFusion \cite{SwinFusion}, CDDFuse \cite{cddfuse}, LapH \cite{laph}, MDA \cite{mda}, ITFuse \cite{itfuse}, CrossFuse \cite{crossfuse}, and EMMA \cite{emma}, and unrolling-based method: LRRNet \cite{lrrnet}. As shown in the figure, the features estimated by the algorithm unrolling-based methods—LRRNet and our proposed FNet—more effectively represent the scenarios in the input image pair compared to the pure DL-based methods. LRRNet estimates low-rank and sparse features from both the images and then combines them to obtain the fused image. However, as illustrated in Fig. \ref{fig_interprtability}-(g), the low-rank and sparse features extracted from the infrared image are very similar. Likewise, the low rank and sprse features from the visible image also share similar structural details. In contrast, the unique features estimated by our FNet are complementary: the unique feature from the visible image captures scene details such as the structural outline of the car and shadows on the road, whereas the unique feature from the infrared image captures the thermal radiation. The common feature captures elements present in both images. For instance, the shadow on the road is not included in the common feature. It is important to note that, although our proposed FNet performs on par with or better than the SOTA methods—such as SwinFusion and CDDFuse (as shown in Table \ref{tab:vif})—our method produces more interpretable features compared to these two approaches.}

\textcolor[rgb]{ 0,  0,  0}{\subsection{Analysis of $\ell _0$ Regularization}
\label{analyze_l0}
In our FNet, we use three LZSC blocks to estimate the unique and common features from the source images of different modalities. In each LZSC block, we use four iteration modules (IMs) to estimate the features. At each IM, the sigmoidal thresholding function $T_{0.1,100,\theta^{(k)}}(\cdot)$ is applied to the features to constrain them to be $\ell _0$ regularized. $T_{0.1,100,\theta^k}(\cdot)$ promotes a sparse solution by nullifying the features with absolute values smaller than the threshold value $\theta^{(k)}$, and imposes negligible penalty for the larger feature values. In this subsection we perform experimental analysis to show that $\ell _0$ regularization is implemented in the three LZSC blocks of FNet. Fig. \ref{fig_threshold} shows the learned threshold values for the four iterations of the three LZSC blocks. The threshold values are positive and decrease monotonically with the iteration numbers, as discussed in subsection \ref{fiht}. Then we compare the histograms of the features before and after applying the thresholding function $T_{0.1,100,\theta^{(k)}}(\cdot)$. The features are taken from the $4^{th}$  iteration module (IM) of the LZSC blocks, for an input image pair from the TNO dataset. As can be seen, $T_{\alpha,\gamma,\theta}(\cdot)$ nullifies the features with absolute values lower than threshold, and imposes negligible penalty for large features. The peak of the histograms at pixel value $0$ shows the number of null features. For the three LZSC blocks estimating $u_1$, $u_2$ and  $c$, the number of null features increase by $84.4\%$  (Fig. \ref{fig_threshold}-(a)), $596.5\%$  (Fig. \ref{fig_threshold}-(b)), and $1068.8\%$  (Fig. \ref{fig_threshold}-(c)), respectively after applying $T_{\alpha,\gamma,\theta^{(4)}}(\cdot)$. Also as seen from the zoomed histograms shown in the respective onsets of the figures, they remain close to the original histograms for pixels with absolute values higher than the thresholds. All these results show that the three LZSC blocks in FNet constrain the features to be $\ell _0$-regularized.}

\begin{table*} [h!]
	\fontsize{7.5}{8}\selectfont
\centering
\caption{mAP@[0.5:0.95] values for object detection on the M3FD dataset, \textcolor[rgb]{ 0,  0,  0}{and mIoU values for semantic segmentation on the MFNet dataset}. We highlight the best and second-best performances in \textcolor{red}{\textbf{red}} and \textcolor{blue}{\textbf{blue}} colors, respectively. High values of mAP@[0.5:0.95] and mIoU are desired.}
\begin{tabular}{r|DEEEEAF|EEDEDEEEEF}
	\toprule
	\multirow{2}[4]{*}{Method} & \multicolumn{7}{c|}{Object Detection, mAP@ [0.5:0.95]} & \multicolumn{10}{c}{\textcolor[rgb]{ 0,  0,  0}{Semantic Segmentation, mIoU}} \\
	\cmidrule{2-18}& \text{People} & \text{Car}   & \text{Truck} & \text{Bus}   & \text{Lamp}  & \text{M.Cycle} & \text{Average} & \textcolor[rgb]{ 0,  0,  0}{\text{Unlbl}} & \textcolor[rgb]{ 0,  0,  0}{\text{Car}}   & \textcolor[rgb]{ 0,  0,  0}{\text{Person}} & \textcolor[rgb]{ 0,  0,  0}{\text{Bike}}  & \textcolor[rgb]{ 0,  0,  0}{\text{Curve}} & \textcolor[rgb]{ 0,  0,  0}{\text{Stop}}  & \textcolor[rgb]{ 0,  0,  0}{\text{G.Rail}} & \textcolor[rgb]{ 0,  0,  0}{\text{Cone}}  & \textcolor[rgb]{ 0,  0,  0}{\text{Bump}}  & \textcolor[rgb]{ 0,  0,  0}{\text{Average}} \\
	\toprule
	Visible & \text{0.473} & \text{0.705} & \textcolor[rgb]{ 0,  0,  1}{\textbf{0.777}} & {0.449} & \text{0.564} & \textcolor[rgb]{ 1,  0,  0}{\textbf{0.633}} & \text{0.600} & \text{0.981} & \text{0.853} & \text{0.639} & \text{0.648} & \text{0.599} & \text{0.676} & \text{0.664} & \text{0.648} & \text{0.806} & \text{0.724} \\
	Infrared & \text{0.553} & \text{0.660} & \text{0.609} & \text{0.470} & \text{0.456} & \text{0.600} & \text{0.573} & \text{0.983} & \text{0.885} & \text{0.746} & \text{0.606} & \text{0.544} & \text{0.570} & \text{0.800} & \text{0.586} & \text{0.754} & \text{0.719} \\
	AUIFNet \cite{auifnet} & \text{0.546} & \text{0.705} & \text{0.729} & \text{0.489} & \textcolor[rgb]{ 1,  0,  0}{\textbf{0.573}} & \text{0.614} & \text{0.610} & \textcolor[rgb]{ 0,  0,  1}{\textbf{0.985}} & \text{0.891} & \text{0.739} & \text{0.627} & \text{0.641} & \text{0.697} & \text{0.636} & \text{0.629} & \textcolor[rgb]{ 0,  0,  1}{\textbf{0.890}} & \text{0.748} \\
	SwinFusion \cite{SwinFusion} & \text{0.546} & \text{0.703} & \text{0.770} & \textcolor[rgb]{ 1,  0,  0}{\textbf{0.503}} & \text{0.558} & \text{0.604} & \text{0.614} & \textcolor[rgb]{ 1,  0,  0}{\textbf{0.986}} & \text{0.895} & \textcolor[rgb]{ 0,  0,  1}{\textbf{0.770}} & \text{0.634} & \text{0.660} & \text{0.695} & \text{0.591} & \text{0.720} & \text{0.723} & \text{0.741} \\
	U2Fusion \cite{U2fusion} & \text{0.552} & \text{0.702} & \text{0.745} & \textcolor[rgb]{ 0,  0,  1}{\textbf{0.501}} & \text{0.566} & \text{0.627} & \text{0.615} & \textcolor[rgb]{ 1,  0,  0}{\textbf{0.986}} & \text{0.893} & \text{0.762} & \textcolor[rgb]{ 1,  0,  0}{\textbf{0.663}} & \textcolor[rgb]{ 1,  0,  0}{\textbf{0.675}} & \textcolor[rgb]{ 0,  0,  1}{\textbf{0.712}} & \text{0.716} & \text{0.705} & \text{0.798} & \text{0.768} \\
	CoCoNet \cite{coconet} & \text{0.536} & \text{0.700} & \text{0.752} & \text{0.523} & \text{0.558} & \text{0.586} & \text{0.609} & \text{0.984} & \text{0.890} & \text{0.741} & \text{0.629} & \text{0.611} & \text{0.599} & \text{0.727} & \text{0.624} & \text{0.727} & \text{0.726} \\
	LapH \cite{laph} & \textcolor[rgb]{ 1,  0,  0}{\textbf{0.564}} & \textcolor[rgb]{ 0,  0,  1}{\textbf{0.706}} & \text{0.775} & \text{0.474} & \text{0.571} & \text{0.621} & \textcolor[rgb]{ 0,  0,  1}{\textbf{0.618}} & \textcolor[rgb]{ 1,  0,  0}{\textbf{0.986}} & \text{0.893} & \text{0.766} & \textcolor[rgb]{ 0,  0,  1}{\textbf{0.659}} & \text{0.657} & \textcolor[rgb]{ 1,  0,  0}{\textbf{0.713}} & \text{0.643} & \text{0.707} & \text{0.860} & \text{0.765} \\
	MURF \cite{murf} & \text{0.547} & \text{0.695} & \text{0.759} & \text{0.499} & \text{0.519} & \text{0.619} & \text{0.607} & \textcolor[rgb]{ 1,  0,  0}{\textbf{0.986}} & \textcolor[rgb]{ 1,  0,  0}{\textbf{0.899}} & \textcolor[rgb]{ 1,  0,  0}{\textbf{0.770}} & \text{0.632} & \text{0.646} & \text{0.675} & \text{0.507} & \textcolor[rgb]{ 1,  0,  0}{\textbf{0.725}} & \text{0.803} & \text{0.738} \\
	LRRNet \cite{lrrnet} & \text{0.534} & \text{0.703} & \text{0.755} & \text{0.480} & \text{0.571} & \text{0.602} & \text{0.608} & \textcolor[rgb]{ 0,  0,  1}{\textbf{0.985}} & \text{0.889} & \text{0.746} & \text{0.638} & \text{0.639} & \text{0.655} & \textcolor[rgb]{ 0,  0,  1}{\textbf{0.745}} & \text{0.649} & \text{0.670} & \text{0.735} \\
	CDDFuse \cite{cddfuse} & \text{0.544} & \text{0.703} & \text{0.773} & \text{0.498} & \text{0.555} & \text{0.601} & \text{0.612} & \textcolor[rgb]{ 0,  0,  1}{\textbf{0.985}} & \text{0.883} & \text{0.758} & \text{0.619} & \text{0.650} & \text{0.630} & \text{0.723} & \text{0.639} & \text{0.843} & \text{0.748} \\
	MDA \cite{mda} & \text{0.480} & \text{0.650} & \text{0.719} & \text{0.405} & \text{0.490} & \text{0.560} & \text{0.551} & \textcolor[rgb]{ 0,  0,  1}{\textbf{0.985}} & \text{0.893} & \text{0.743} & \text{0.621} & \text{0.646} & \text{0.698} & \text{0.737} & \text{0.708} & \text{0.429} & \text{0.718} \\
	ITFuse \cite{itfuse} & \text{0.539} & \text{0.700} & \text{0.733} & \text{0.492} & \text{0.557} & \text{0.626} & \text{0.608} & \textcolor[rgb]{ 0,  0,  1}{\textbf{0.985}} & \text{0.889} & \text{0.734} & \text{0.635} & \text{0.640} & \text{0.668} & \text{0.692} & \text{0.660} & \text{0.804} & \text{0.745} \\
	CrossFuse \cite{crossfuse} & \text{0.527} & \text{0.705} & \textcolor[rgb]{ 1,  0,  0}{\textbf{0.794}} & \text{0.470} & \text{0.562} & \text{0.626} & \text{0.614} & \textcolor[rgb]{ 1,  0,  0}{\textbf{0.986}} & \text{0.897} & \text{0.752} & \text{0.642} & \text{0.658} & \text{0.701} & \textcolor[rgb]{ 1,  0,  0}{\textbf{0.754}} & \text{0.708} & \text{0.834} & \textcolor[rgb]{ 0,  0,  1}{\textbf{0.770}} \\
	EMMA \cite{emma} & \text{0.541} & \text{0.703} & \text{0.767} & \text{0.498} & \text{0.558} & \text{0.621} & \text{0.615} & \textcolor[rgb]{ 0,  0,  1}{\textbf{0.985}} & \text{0.877} & \text{0.758} & \text{0.623} & \text{0.648} & \text{0.650} & \text{0.716} & \text{0.651} & \text{0.869} & \text{0.753} \\
	FNet (Ours) & \textcolor[rgb]{ 0,  0,  1}{\textbf{0.554}} & \textcolor[rgb]{ 1,  0,  0}{\textbf{0.707}} & \text{0.755} & \text{0.500} & \textcolor[rgb]{ 0,  0,  1}{\textbf{0.572}} & \textcolor[rgb]{ 0,  0,  1}{\textbf{0.631}} & \textcolor[rgb]{ 1,  0,  0}{\textbf{0.620}} & \textcolor[rgb]{ 1,  0,  0}{\textbf{0.986}} & \textcolor[rgb]{ 0,  0,  1}{\textbf{0.898}} & \textcolor[rgb]{ 0,  0,  1}{\textbf{0.769}} & \text{0.643} & \textcolor[rgb]{ 0,  0,  1}{\textbf{0.668}} & \text{0.704} & \text{0.726} & \textcolor[rgb]{ 0,  0,  1}{\textbf{0.724}} & \textcolor[rgb]{ 1,  0,  0}{\textbf{0.901}} & \textcolor[rgb]{ 1,  0,  0}{\textbf{0.780}} \\
	\bottomrule
\end{tabular}

\label{tab:object}
\end{table*}
\begin{figure*}[b!] 
    \centering
  \includegraphics[width=0.8\linewidth]{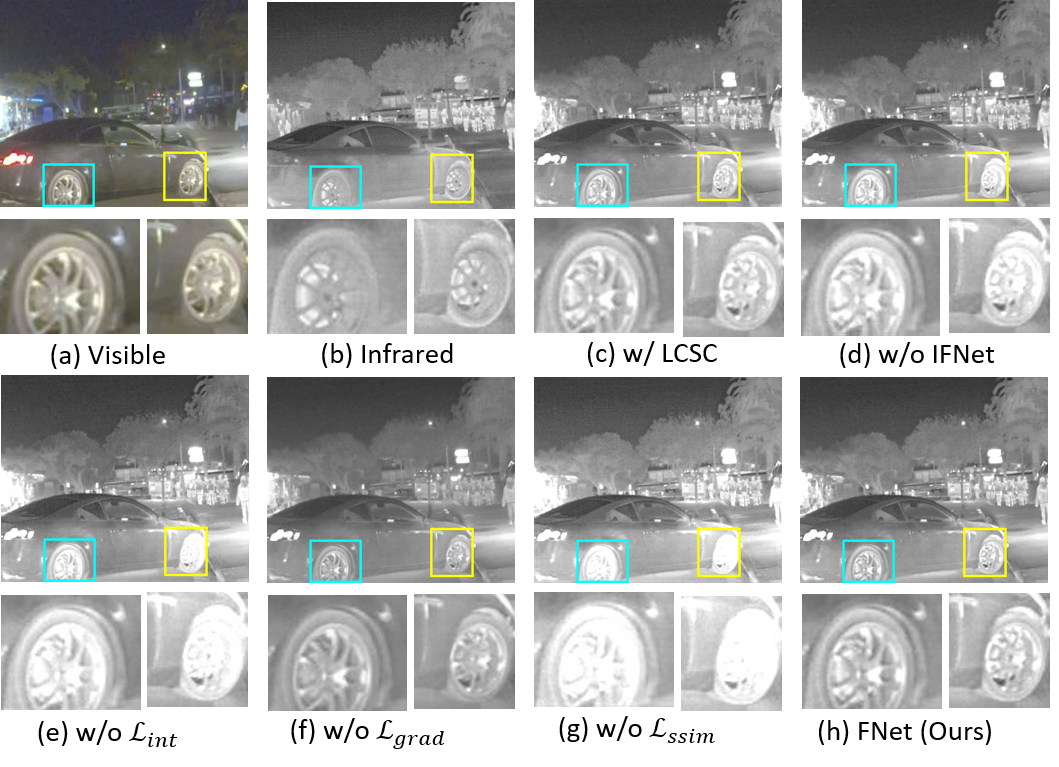}
  
  \caption{Visual comparison of the ablation experiments for the VIS-IR task on the RoadScene dataset.}
  \label{fig11} 
\end{figure*}

\begin{table*}[h!]
	\fontsize{7.5}{8}\selectfont
	\centering
	\caption{Comparison of results on ablation experiments (AE). We highlight the best and second-best performances in \textcolor{red}{\textbf{red}} and \textcolor{blue}{\textbf{blue}} colors, respectively. $\uparrow$ means high value, and $\downarrow$ means low value desired.}
    \begin{tabular}{P|r|EEEEEEEEEEEEEED|E}
	\toprule
	\multicolumn{1}{c|}{Tasks (Datasets)} & AE    & AE1 & \textcolor{black}{AE2} & \textcolor{black}{AE3} & \textcolor{black}{AE4} & \textcolor{black}{AE5} & \textcolor{black}{AE6} & AE7 & AE8 & AE9 & AE10 & AE11 & \textcolor{black}{AE12} & AE13 & AE14 & AE15 & \multicolumn{1}{r}{FNet(Ours)} \\
	\midrule
\multicolumn{1}{c|}{\multirow{6}[2]{*}{\makecell{VIS-IR \\ (TNO)}}} & MI ↑  & 2.35  & 2.46  & 2.38  & 2.45  & 2.42  & \textcolor[rgb]{ 0,  0,  1}{\textbf{2.55}} & 2.39  & 2.45  & 2.39  & 2.37  & 2.22  & 2.47  & 1.83  & 2.38  & 1.67  & \textcolor[rgb]{ 1,  0,  0}{\textbf{2.57}} \\
	& CE ↓  & 2.58  & 2.72  & 2.81  & 2.82  & 2.71  & 2.65  & 2.55  & 2.49  & 2.69  & 2.66  & 2.75  & 2.72  & 3.01  & \textcolor[rgb]{ 0,  0,  1}{\textbf{2.48}} & 3.87  & \textcolor[rgb]{ 1,  0,  0}{\textbf{2.20}} \\
	& VIF ↑ & 0.78  & \textcolor[rgb]{ 0,  0,  1}{\textbf{0.79}} & \textcolor[rgb]{ 0,  0,  1}{\textbf{0.79}} & 0.78  & 0.78  & \textcolor[rgb]{ 1,  0,  0}{\textbf{0.80}} & 0.78  & \textcolor[rgb]{ 0,  0,  1}{\textbf{0.79}} & 0.77  & 0.78  & 0.76  & \textcolor[rgb]{ 0,  0,  1}{\textbf{0.79}} & 0.70  & 0.74  & 0.64  & \textcolor[rgb]{ 1,  0,  0}{\textbf{0.80}} \\
	& Qabf ↑ & 0.55  & 0.55  & \textcolor[rgb]{ 0,  0,  1}{\textbf{0.56}} & 0.55  & 0.55  & \textcolor[rgb]{ 0,  0,  1}{\textbf{0.56}} & 0.55  & 0.55  & 0.55  & 0.55  & 0.54  & \textcolor[rgb]{ 0,  0,  1}{\textbf{0.56}} & 0.52  & 0.51  & 0.48  & \textcolor[rgb]{ 1,  0,  0}{\textbf{0.57}} \\
	& Qcb ↑ & \textcolor[rgb]{ 0,  0,  1}{\textbf{0.51}} & \textcolor[rgb]{ 1,  0,  0}{\textbf{0.52}} & \textcolor[rgb]{ 1,  0,  0}{\textbf{0.52}} & \textcolor[rgb]{ 1,  0,  0}{\textbf{0.52}} & \textcolor[rgb]{ 1,  0,  0}{\textbf{0.52}} & \textcolor[rgb]{ 1,  0,  0}{\textbf{0.52}} & \textcolor[rgb]{ 0,  0,  1}{\textbf{0.51}} & \textcolor[rgb]{ 1,  0,  0}{\textbf{0.52}} & \textcolor[rgb]{ 1,  0,  0}{\textbf{0.52}} & \textcolor[rgb]{ 1,  0,  0}{\textbf{0.52}} & \textcolor[rgb]{ 0,  0,  1}{\textbf{0.51}} & \textcolor[rgb]{ 1,  0,  0}{\textbf{0.52}} & 0.49  & 0.50  & 0.47  & \textcolor[rgb]{ 1,  0,  0}{\textbf{0.52}} \\
	& SSIM ↑ & \textcolor[rgb]{ 1,  0,  0}{\textbf{1.05}} & \textcolor[rgb]{ 0,  0,  1}{\textbf{1.04}} & \textcolor[rgb]{ 0,  0,  1}{\textbf{1.04}} & \textcolor[rgb]{ 0,  0,  1}{\textbf{1.04}} & \textcolor[rgb]{ 0,  0,  1}{\textbf{1.04}} & \textcolor[rgb]{ 0,  0,  1}{\textbf{1.04}} & \textcolor[rgb]{ 0,  0,  1}{\textbf{1.04}} & \textcolor[rgb]{ 1,  0,  0}{\textbf{1.05}} & \textcolor[rgb]{ 1,  0,  0}{\textbf{1.05}} & \textcolor[rgb]{ 0,  0,  1}{\textbf{1.04}} & \textcolor[rgb]{ 0,  0,  1}{\textbf{1.04}} & \textcolor[rgb]{ 1,  0,  0}{\textbf{1.05}} & 1.03  & 1.03  & 1.01  & \textcolor[rgb]{ 1,  0,  0}{\textbf{1.05}} \\
	\midrule
	\multicolumn{1}{c|}{\multirow{6}[2]{*}{\makecell{VIS-IR \\ (Road- \\ Scene)}}} & MI ↑  & 2.38  & 2.45  & 2.39  & 2.50  & \textcolor[rgb]{ 0,  0,  1}{\textbf{2.51}} & \textcolor[rgb]{ 1,  0,  0}{\textbf{2.56}} & 2.40  & 2.43  & 2.45  & 2.43  & 2.34  & 2.47  & 1.99  & 2.50  & 1.98  & \textcolor[rgb]{ 1,  0,  0}{\textbf{2.56}} \\
	& CE ↓  & 2.03  & 1.87  & 1.99  & 1.79  & 1.77  & \textcolor[rgb]{ 0,  0,  1}{\textbf{1.74}} & 1.93  & 1.90  & 1.87  & 1.93  & 2.03  & 1.81  & 2.99  & \textcolor[rgb]{ 1,  0,  0}{\textbf{1.44}} & 3.60  & 1.81 \\
	& VIF ↑ & 0.67  & 0.67  & 0.68  & 0.68  & \textcolor[rgb]{ 0,  0,  1}{\textbf{0.69}} & \textcolor[rgb]{ 0,  0,  1}{\textbf{0.69}} & 0.67  & 0.68  & 0.68  & 0.68  & 0.66  & \textcolor[rgb]{ 0,  0,  1}{\textbf{0.69}} & 0.62  & 0.67  & 0.62  & \textcolor[rgb]{ 1,  0,  0}{\textbf{0.70}} \\
	& Qabf ↑ & 0.49  & \textcolor[rgb]{ 0,  0,  1}{\textbf{0.50}} & \textcolor[rgb]{ 0,  0,  1}{\textbf{0.50}} & \textcolor[rgb]{ 0,  0,  1}{\textbf{0.50}} & \textcolor[rgb]{ 1,  0,  0}{\textbf{0.51}} & \textcolor[rgb]{ 1,  0,  0}{\textbf{0.51}} & \textcolor[rgb]{ 0,  0,  1}{\textbf{0.50}} & \textcolor[rgb]{ 1,  0,  0}{\textbf{0.51}} & \textcolor[rgb]{ 1,  0,  0}{\textbf{0.51}} & \textcolor[rgb]{ 0,  0,  1}{\textbf{0.50}} & \textcolor[rgb]{ 0,  0,  1}{\textbf{0.50}} & \textcolor[rgb]{ 1,  0,  0}{\textbf{0.51}} & \textcolor[rgb]{ 0,  0,  1}{\textbf{0.50}} & 0.42  & 0.49  & \textcolor[rgb]{ 1,  0,  0}{\textbf{0.51}} \\
	& Qcb ↑ & 0.46  & 0.47  & 0.46  & 0.47  & \textcolor[rgb]{ 1,  0,  0}{\textbf{0.49}} & \textcolor[rgb]{ 0,  0,  1}{\textbf{0.48}} & 0.47  & 0.47  & \textcolor[rgb]{ 0,  0,  1}{\textbf{0.48}} & 0.47  & 0.46  & 0.47  & 0.44  & \textcolor[rgb]{ 1,  0,  0}{\textbf{0.49}} & 0.45  & \textcolor[rgb]{ 0,  0,  1}{\textbf{0.48}} \\
	& SSIM ↑ & \textcolor[rgb]{ 1,  0,  0}{\textbf{1.01}} & \textcolor[rgb]{ 1,  0,  0}{\textbf{1.01}} & \textcolor[rgb]{ 1,  0,  0}{\textbf{1.01}} & \textcolor[rgb]{ 0,  0,  1}{\textbf{1.00}} & \textcolor[rgb]{ 1,  0,  0}{\textbf{1.01}} & \textcolor[rgb]{ 1,  0,  0}{\textbf{1.01}} & 0.99  & 0.99  & \textcolor[rgb]{ 1,  0,  0}{\textbf{1.01}} & \textcolor[rgb]{ 0,  0,  1}{\textbf{1.00}} & \textcolor[rgb]{ 1,  0,  0}{\textbf{1.01}} & \textcolor[rgb]{ 1,  0,  0}{\textbf{1.01}} & 0.97  & 0.96  & 0.93  & \textcolor[rgb]{ 1,  0,  0}{\textbf{1.01}} \\
	\midrule
	\multicolumn{1}{c|}{\multirow{6}[2]{*}{\makecell{VIS-IR \\ (CATS)}}} & MI ↑  & 2.60  & 2.69  & 2.66  & 2.68  & 2.68  & \textcolor[rgb]{ 0,  0,  1}{\textbf{2.73}} & 2.60  & 2.66  & 2.65  & 2.64  & 2.47  & 2.70  & 2.32  & 2.56  & 2.20  & \textcolor[rgb]{ 1,  0,  0}{\textbf{2.75}} \\
	& CE ↓  & 3.00  & 2.86  & 2.80  & 2.77  & 2.76  & \textcolor[rgb]{ 1,  0,  0}{\textbf{2.69}} & 2.84  & 2.93  & 2.84  & 2.87  & 2.84  & 2.79  & 3.18  & 2.86  & 3.09  & \textcolor[rgb]{ 0,  0,  1}{\textbf{2.75}} \\
	& VIF ↑ & 0.84  & 0.85  & \textcolor[rgb]{ 1,  0,  0}{\textbf{0.89}} & 0.87  & \textcolor[rgb]{ 0,  0,  1}{\textbf{0.88}} & \textcolor[rgb]{ 0,  0,  1}{\textbf{0.88}} & 0.85  & 0.86  & \textcolor[rgb]{ 0,  0,  1}{\textbf{0.88}} & 0.87  & 0.83  & \textcolor[rgb]{ 0,  0,  1}{\textbf{0.88}} & 0.87  & 0.81  & 0.81  & \textcolor[rgb]{ 1,  0,  0}{\textbf{0.89}} \\
	& Qabf ↑ & \textcolor[rgb]{ 0,  0,  1}{\textbf{0.67}} & \textcolor[rgb]{ 0,  0,  1}{\textbf{0.67}} & \textcolor[rgb]{ 1,  0,  0}{\textbf{0.68}} & \textcolor[rgb]{ 0,  0,  1}{\textbf{0.67}} & \textcolor[rgb]{ 0,  0,  1}{\textbf{0.67}} & \textcolor[rgb]{ 0,  0,  1}{\textbf{0.67}} & \textcolor[rgb]{ 0,  0,  1}{\textbf{0.67}} & \textcolor[rgb]{ 0,  0,  1}{\textbf{0.67}} & \textcolor[rgb]{ 1,  0,  0}{\textbf{0.68}} & \textcolor[rgb]{ 0,  0,  1}{\textbf{0.67}} & \textcolor[rgb]{ 0,  0,  1}{\textbf{0.67}} & \textcolor[rgb]{ 1,  0,  0}{\textbf{0.68}} & \textcolor[rgb]{ 0,  0,  1}{\textbf{0.67}} & 0.65  & 0.61  & \textcolor[rgb]{ 1,  0,  0}{\textbf{0.68}} \\
	& Qcb ↑ & 0.53  & \textcolor[rgb]{ 0,  0,  1}{\textbf{0.54}} & \textcolor[rgb]{ 0,  0,  1}{\textbf{0.54}} & \textcolor[rgb]{ 0,  0,  1}{\textbf{0.54}} & \textcolor[rgb]{ 0,  0,  1}{\textbf{0.54}} & \textcolor[rgb]{ 0,  0,  1}{\textbf{0.54}} & 0.53  & \textcolor[rgb]{ 0,  0,  1}{\textbf{0.54}} & \textcolor[rgb]{ 0,  0,  1}{\textbf{0.54}} & \textcolor[rgb]{ 0,  0,  1}{\textbf{0.54}} & 0.53  & \textcolor[rgb]{ 0,  0,  1}{\textbf{0.54}} & \textcolor[rgb]{ 1,  0,  0}{\textbf{0.55}} & \textcolor[rgb]{ 0,  0,  1}{\textbf{0.54}} & 0.49  & \textcolor[rgb]{ 0,  0,  1}{\textbf{0.54}} \\
	& SSIM ↑ & 0.93  & \textcolor[rgb]{ 0,  0,  1}{\textbf{0.94}} & \textcolor[rgb]{ 0,  0,  1}{\textbf{0.94}} & \textcolor[rgb]{ 0,  0,  1}{\textbf{0.94}} & \textcolor[rgb]{ 1,  0,  0}{\textbf{0.95}} & \textcolor[rgb]{ 0,  0,  1}{\textbf{0.94}} & \textcolor[rgb]{ 0,  0,  1}{\textbf{0.94}} & 0.93  & \textcolor[rgb]{ 1,  0,  0}{\textbf{0.95}} & \textcolor[rgb]{ 0,  0,  1}{\textbf{0.94}} & 0.91  & \textcolor[rgb]{ 1,  0,  0}{\textbf{0.95}} & \textcolor[rgb]{ 1,  0,  0}{\textbf{0.95}} & 0.93  & 0.93  & \textcolor[rgb]{ 1,  0,  0}{\textbf{0.95}} \\
	\midrule
	\multicolumn{1}{c|}{\multirow{6}[2]{*}{\makecell{VIS-IR \\ (KAIST)}}} & MI ↑  & 2.88  & 2.89  & 2.90  & 2.95  & \textcolor[rgb]{ 0,  0,  1}{\textbf{2.98}} & 2.97  & 2.83  & 2.93  & 2.85  & 2.87  & 2.78  & 2.96  & 2.42  & 2.79  & 2.43  & \textcolor[rgb]{ 1,  0,  0}{\textbf{2.99}} \\
	& CE ↓  & 2.68  & 2.71  & 2.65  & 2.39  & 2.63  & 2.62  & 2.67  & 2.48  & 2.52  & 2.65  & 2.50  & \textcolor[rgb]{ 0,  0,  1}{\textbf{2.37}} & \textcolor[rgb]{ 1,  0,  0}{\textbf{2.17}} & 2.55  & 2.88  & 2.63 \\
	& VIF ↑ & 0.81  & 0.81  & 0.82  & \textcolor[rgb]{ 0,  0,  1}{\textbf{0.83}} & \textcolor[rgb]{ 1,  0,  0}{\textbf{0.84}} & \textcolor[rgb]{ 1,  0,  0}{\textbf{0.84}} & 0.80  & 0.82  & 0.81  & 0.81  & 0.80  & \textcolor[rgb]{ 1,  0,  0}{\textbf{0.84}} & \textcolor[rgb]{ 0,  0,  1}{\textbf{0.83}} & 0.78  & 0.76  & \textcolor[rgb]{ 1,  0,  0}{\textbf{0.84}} \\
	& Qabf ↑ & \textcolor[rgb]{ 1,  0,  0}{\textbf{0.63}} & \textcolor[rgb]{ 1,  0,  0}{\textbf{0.63}} & \textcolor[rgb]{ 1,  0,  0}{\textbf{0.63}} & \textcolor[rgb]{ 1,  0,  0}{\textbf{0.63}} & \textcolor[rgb]{ 1,  0,  0}{\textbf{0.63}} & \textcolor[rgb]{ 1,  0,  0}{\textbf{0.63}} & \textcolor[rgb]{ 0,  0,  1}{\textbf{0.62}} & \textcolor[rgb]{ 1,  0,  0}{\textbf{0.63}} & \textcolor[rgb]{ 0,  0,  1}{\textbf{0.62}} & \textcolor[rgb]{ 0,  0,  1}{\textbf{0.62}} & \textcolor[rgb]{ 0,  0,  1}{\textbf{0.62}} & \textcolor[rgb]{ 1,  0,  0}{\textbf{0.63}} & \textcolor[rgb]{ 1,  0,  0}{\textbf{0.63}} & 0.60  & 0.57  & \textcolor[rgb]{ 1,  0,  0}{\textbf{0.63}} \\
	& Qcb ↑ & 0.51  & 0.51  & 0.51  & 0.50  & 0.50  & 0.50  & 0.51  & 0.51  & 0.50  & 0.50  & 0.51  & \textcolor[rgb]{ 0,  0,  1}{\textbf{0.52}} & \textcolor[rgb]{ 1,  0,  0}{\textbf{0.53}} & 0.47  & 0.46  & 0.50 \\
	& SSIM ↑ & 0.95  & 0.95  & 0.95  & 0.95  & \textcolor[rgb]{ 0,  0,  1}{\textbf{0.96}} & 0.95  & 0.95  & \textcolor[rgb]{ 0,  0,  1}{\textbf{0.96}} & 0.95  & \textcolor[rgb]{ 0,  0,  1}{\textbf{0.96}} & 0.94  & \textcolor[rgb]{ 0,  0,  1}{\textbf{0.96}} & \textcolor[rgb]{ 1,  0,  0}{\textbf{0.97}} & 0.94  & \textcolor[rgb]{ 0,  0,  1}{\textbf{0.96}} & \textcolor[rgb]{ 1,  0,  0}{\textbf{0.97}} \\
	\midrule
	\multicolumn{1}{c|}{\multirow{6}[2]{*}{\makecell{VIS-NIR \\ (RGB-NIR \\ Scene)}}} & MI ↑  & 2.07  & 2.07  & 2.05  & 2.05  & 2.00  & 2.07  & 2.08  & 2.06  & 2.06  & 2.04  & 2.02  & 2.03  & 2.08  & \textcolor[rgb]{ 1,  0,  0}{\textbf{2.20}} & 2.01  & \textcolor[rgb]{ 0,  0,  1}{\textbf{2.09}} \\
	& CE ↓  & 1.82  & \textcolor[rgb]{ 0,  0,  1}{\textbf{1.77}} & \textcolor[rgb]{ 1,  0,  0}{\textbf{1.75}} & 1.83  & 1.90  & 1.85  & 1.87  & 1.86  & 1.86  & 1.81  & 1.83  & 1.86  & 1.87  & 1.85  & 2.60  & 1.87 \\
	& VIF ↑ & 0.62  & 0.63  & \textcolor[rgb]{ 0,  0,  1}{\textbf{0.64}} & 0.63  & 0.62  & 0.63  & 0.62  & 0.63  & 0.63  & 0.62  & 0.61  & 0.63  & 0.62  & \textcolor[rgb]{ 1,  0,  0}{\textbf{0.65}} & 0.59  & 0.63 \\
	& Qabf ↑ & 0.44  & \textcolor[rgb]{ 0,  0,  1}{\textbf{0.45}} & 0.44  & \textcolor[rgb]{ 0,  0,  1}{\textbf{0.45}} & 0.44  & \textcolor[rgb]{ 0,  0,  1}{\textbf{0.45}} & 0.43  & \textcolor[rgb]{ 0,  0,  1}{\textbf{0.45}} & \textcolor[rgb]{ 0,  0,  1}{\textbf{0.45}} & 0.44  & 0.43  & \textcolor[rgb]{ 0,  0,  1}{\textbf{0.45}} & \textcolor[rgb]{ 0,  0,  1}{\textbf{0.45}} & \textcolor[rgb]{ 1,  0,  0}{\textbf{0.47}} & 0.41  & \textcolor[rgb]{ 0,  0,  1}{\textbf{0.45}} \\
	& Qcb ↑ & \textcolor[rgb]{ 1,  0,  0}{\textbf{0.51}} & \textcolor[rgb]{ 0,  0,  1}{\textbf{0.50}} & \textcolor[rgb]{ 1,  0,  0}{\textbf{0.51}} & \textcolor[rgb]{ 0,  0,  1}{\textbf{0.50}} & 0.49  & \textcolor[rgb]{ 0,  0,  1}{\textbf{0.50}} & \textcolor[rgb]{ 1,  0,  0}{\textbf{0.51}} & \textcolor[rgb]{ 1,  0,  0}{\textbf{0.51}} & \textcolor[rgb]{ 0,  0,  1}{\textbf{0.50}} & \textcolor[rgb]{ 0,  0,  1}{\textbf{0.50}} & \textcolor[rgb]{ 0,  0,  1}{\textbf{0.50}} & \textcolor[rgb]{ 0,  0,  1}{\textbf{0.50}} & 0.49  & 0.49  & 0.46  & \textcolor[rgb]{ 1,  0,  0}{\textbf{0.51}} \\
	& SSIM ↑ & \textcolor[rgb]{ 0,  0,  1}{\textbf{1.20}} & \textcolor[rgb]{ 0,  0,  1}{\textbf{1.20}} & 1.19  & \textcolor[rgb]{ 0,  0,  1}{\textbf{1.20}} & 1.19  & \textcolor[rgb]{ 0,  0,  1}{\textbf{1.20}} & 1.19  & \textcolor[rgb]{ 0,  0,  1}{\textbf{1.20}} & \textcolor[rgb]{ 1,  0,  0}{\textbf{1.21}} & 1.19  & 1.18  & \textcolor[rgb]{ 0,  0,  1}{\textbf{1.20}} & 1.19  & \textcolor[rgb]{ 1,  0,  0}{\textbf{1.21}} & 1.14  & \textcolor[rgb]{ 0,  0,  1}{\textbf{1.20}} \\
	\midrule
	\multicolumn{1}{c|}{\multirow{6}[2]{*}{\makecell{CT-MRI \\ (Harvard \\ medical)}}} & MI ↑  & 2.22  & 2.22  & 2.15  & 2.21  & 2.22  & 2.22  & 2.22  & 2.18  & 2.19  & 2.20  & 2.18  & 2.20  & 2.14  & \textcolor[rgb]{ 1,  0,  0}{\textbf{2.30}} & 2.12  & \textcolor[rgb]{ 0,  0,  1}{\textbf{2.23}} \\
	& CE ↓  & 0.23  & \textcolor[rgb]{ 0,  0,  1}{\textbf{0.22}} & \textcolor[rgb]{ 1,  0,  0}{\textbf{0.20}} & \textcolor[rgb]{ 1,  0,  0}{\textbf{0.20}} & \textcolor[rgb]{ 0,  0,  1}{\textbf{0.22}} & \textcolor[rgb]{ 1,  0,  0}{\textbf{0.20}} & 0.23  & \textcolor[rgb]{ 0,  0,  1}{\textbf{0.22}} & 0.23  & 0.23  & 0.23  & \textcolor[rgb]{ 1,  0,  0}{\textbf{0.20}} & 0.41  & 0.23  & 0.44  & \textcolor[rgb]{ 1,  0,  0}{\textbf{0.20}} \\
	& VIF ↑ & 0.55  & 0.56  & 0.52  & 0.56  & \textcolor[rgb]{ 0,  0,  1}{\textbf{0.58}} & 0.57  & 0.57  & 0.54  & 0.54  & 0.55  & 0.53  & 0.56  & 0.50  & \textcolor[rgb]{ 1,  0,  0}{\textbf{0.63}} & 0.50  & 0.57 \\
	& Qabf ↑ & \textcolor[rgb]{ 0,  0,  1}{\textbf{0.58}} & \textcolor[rgb]{ 0,  0,  1}{\textbf{0.58}} & 0.57  & \textcolor[rgb]{ 0,  0,  1}{\textbf{0.58}} & \textcolor[rgb]{ 0,  0,  1}{\textbf{0.58}} & \textcolor[rgb]{ 0,  0,  1}{\textbf{0.58}} & \textcolor[rgb]{ 0,  0,  1}{\textbf{0.58}} & 0.57  & \textcolor[rgb]{ 0,  0,  1}{\textbf{0.58}} & \textcolor[rgb]{ 0,  0,  1}{\textbf{0.58}} & \textcolor[rgb]{ 0,  0,  1}{\textbf{0.58}} & \textcolor[rgb]{ 0,  0,  1}{\textbf{0.58}} & \textcolor[rgb]{ 0,  0,  1}{\textbf{0.58}} & 0.49  & 0.57  & \textcolor[rgb]{ 1,  0,  0}{\textbf{0.59}} \\
	& Qcb ↑ & 0.66  & \textcolor[rgb]{ 0,  0,  1}{\textbf{0.67}} & 0.66  & \textcolor[rgb]{ 1,  0,  0}{\textbf{0.68}} & \textcolor[rgb]{ 1,  0,  0}{\textbf{0.68}} & \textcolor[rgb]{ 1,  0,  0}{\textbf{0.68}} & \textcolor[rgb]{ 1,  0,  0}{\textbf{0.68}} & \textcolor[rgb]{ 0,  0,  1}{\textbf{0.67}} & \textcolor[rgb]{ 0,  0,  1}{\textbf{0.67}} & \textcolor[rgb]{ 1,  0,  0}{\textbf{0.68}} & \textcolor[rgb]{ 0,  0,  1}{\textbf{0.67}} & \textcolor[rgb]{ 1,  0,  0}{\textbf{0.68}} & \textcolor[rgb]{ 1,  0,  0}{\textbf{0.68}} & \textcolor[rgb]{ 0,  0,  1}{\textbf{0.67}} & \textcolor[rgb]{ 0,  0,  1}{\textbf{0.67}} & \textcolor[rgb]{ 1,  0,  0}{\textbf{0.68}} \\
	& SSIM ↑ & 1.28  & 1.32  & 1.30  & \textcolor[rgb]{ 1,  0,  0}{\textbf{1.34}} & \textcolor[rgb]{ 1,  0,  0}{\textbf{1.34}} & \textcolor[rgb]{ 0,  0,  1}{\textbf{1.33}} & \textcolor[rgb]{ 1,  0,  0}{\textbf{1.34}} & 1.31  & \textcolor[rgb]{ 0,  0,  1}{\textbf{1.33}} & \textcolor[rgb]{ 0,  0,  1}{\textbf{1.33}} & 1.32  & \textcolor[rgb]{ 1,  0,  0}{\textbf{1.34}} & \textcolor[rgb]{ 1,  0,  0}{\textbf{1.34}} & 1.29  & 1.28  & \textcolor[rgb]{ 1,  0,  0}{\textbf{1.34}} \\
	\midrule
	\multicolumn{1}{c|}{\multirow{6}[2]{*}{\makecell{PET-MRI \\ (Harvard \\ medical)}}} & MI ↑  & 1.88  & 1.94  & 1.91  & 1.93  & \textcolor[rgb]{ 0,  0,  1}{\textbf{1.96}} & \textcolor[rgb]{ 0,  0,  1}{\textbf{1.96}} & 1.92  & 1.95  & 1.94  & 1.93  & 1.91  & \textcolor[rgb]{ 1,  0,  0}{\textbf{1.97}} & 1.84  & 1.89  & 1.84  & \textcolor[rgb]{ 1,  0,  0}{\textbf{1.97}} \\
	& CE ↓  & 0.80  & 0.79  & 0.78  & 0.78  & 0.79  & 0.78  & 0.78  & 0.78  & 0.78  & \textcolor[rgb]{ 0,  0,  1}{\textbf{0.77}} & 0.78  & 0.78  & 0.83  & \textcolor[rgb]{ 1,  0,  0}{\textbf{0.76}} & 0.84  & 0.78 \\
	& VIF ↑ & 0.66  & 0.70  & 0.67  & 0.70  & \textcolor[rgb]{ 0,  0,  1}{\textbf{0.71}} & \textcolor[rgb]{ 1,  0,  0}{\textbf{0.72}} & \textcolor[rgb]{ 0,  0,  1}{\textbf{0.71}} & 0.70  & 0.68  & 0.68  & 0.67  & 0.70  & 0.59  & 0.70  & 0.58  & \textcolor[rgb]{ 0,  0,  1}{\textbf{0.71}} \\
	& Qabf ↑ & 0.64  & \textcolor[rgb]{ 1,  0,  0}{\textbf{0.66}} & \textcolor[rgb]{ 0,  0,  1}{\textbf{0.65}} & \textcolor[rgb]{ 1,  0,  0}{\textbf{0.66}} & \textcolor[rgb]{ 1,  0,  0}{\textbf{0.66}} & \textcolor[rgb]{ 1,  0,  0}{\textbf{0.66}} & \textcolor[rgb]{ 0,  0,  1}{\textbf{0.65}} & \textcolor[rgb]{ 1,  0,  0}{\textbf{0.66}} & \textcolor[rgb]{ 0,  0,  1}{\textbf{0.65}} & \textcolor[rgb]{ 0,  0,  1}{\textbf{0.65}} & \textcolor[rgb]{ 0,  0,  1}{\textbf{0.65}} & \textcolor[rgb]{ 1,  0,  0}{\textbf{0.66}} & 0.63  & 0.41  & 0.63  & \textcolor[rgb]{ 1,  0,  0}{\textbf{0.66}} \\
	& Qcb ↑ & \textcolor[rgb]{ 1,  0,  0}{\textbf{0.71}} & \textcolor[rgb]{ 0,  0,  1}{\textbf{0.70}} & \textcolor[rgb]{ 1,  0,  0}{\textbf{0.71}} & \textcolor[rgb]{ 1,  0,  0}{\textbf{0.71}} & \textcolor[rgb]{ 1,  0,  0}{\textbf{0.71}} & \textcolor[rgb]{ 1,  0,  0}{\textbf{0.71}} & \textcolor[rgb]{ 0,  0,  1}{\textbf{0.70}} & \textcolor[rgb]{ 1,  0,  0}{\textbf{0.71}} & \textcolor[rgb]{ 1,  0,  0}{\textbf{0.71}} & \textcolor[rgb]{ 1,  0,  0}{\textbf{0.71}} & \textcolor[rgb]{ 1,  0,  0}{\textbf{0.71}} & \textcolor[rgb]{ 1,  0,  0}{\textbf{0.71}} & \textcolor[rgb]{ 1,  0,  0}{\textbf{0.71}} & 0.69  & \textcolor[rgb]{ 1,  0,  0}{\textbf{0.71}} & \textcolor[rgb]{ 1,  0,  0}{\textbf{0.71}} \\
	& SSIM ↑ & 1.40  & \textcolor[rgb]{ 0,  0,  1}{\textbf{1.49}} & \textcolor[rgb]{ 0,  0,  1}{\textbf{1.49}} & \textcolor[rgb]{ 1,  0,  0}{\textbf{1.50}} & \textcolor[rgb]{ 1,  0,  0}{\textbf{1.50}} & \textcolor[rgb]{ 1,  0,  0}{\textbf{1.50}} & \textcolor[rgb]{ 0,  0,  1}{\textbf{1.49}} & \textcolor[rgb]{ 1,  0,  0}{\textbf{1.50}} & \textcolor[rgb]{ 1,  0,  0}{\textbf{1.50}} & \textcolor[rgb]{ 0,  0,  1}{\textbf{1.49}} & \textcolor[rgb]{ 0,  0,  1}{\textbf{1.49}} & \textcolor[rgb]{ 1,  0,  0}{\textbf{1.50}} & 1.48  & 1.38  & 1.48  & \textcolor[rgb]{ 1,  0,  0}{\textbf{1.50}} \\
	\midrule
	\multicolumn{1}{c|}{\multirow{6}[2]{*}{\makecell{SPECT-MRI \\ (Harvard \\ medical)}}} & MI ↑  & 1.87  & 1.88  & 1.87  & 1.86  & \textcolor[rgb]{ 0,  0,  1}{\textbf{1.90}} & \textcolor[rgb]{ 1,  0,  0}{\textbf{1.91}} & 1.88  & 1.88  & 1.86  & 1.87  & 1.86  & 1.89  & 1.78  & 1.83  & 1.85  & \textcolor[rgb]{ 1,  0,  0}{\textbf{1.91}} \\
	& CE ↓  & 1.13  & 1.18  & 1.14  & \textcolor[rgb]{ 0,  0,  1}{\textbf{1.12}} & \textcolor[rgb]{ 0,  0,  1}{\textbf{1.12}} & \textcolor[rgb]{ 1,  0,  0}{\textbf{1.11}} & 1.15  & 1.13  & 1.13  & 1.13  & 1.13  & \textcolor[rgb]{ 1,  0,  0}{\textbf{1.11}} & 1.36  & 1.15  & 1.30  & \textcolor[rgb]{ 1,  0,  0}{\textbf{1.11}} \\
	& VIF ↑ & 0.62  & 0.62  & 0.62  & \textcolor[rgb]{ 0,  0,  1}{\textbf{0.63}} & 0.62  & \textcolor[rgb]{ 1,  0,  0}{\textbf{0.64}} & 0.62  & \textcolor[rgb]{ 0,  0,  1}{\textbf{0.63}} & 0.62  & 0.62  & 0.62  & \textcolor[rgb]{ 0,  0,  1}{\textbf{0.63}} & 0.59  & 0.57  & \textcolor[rgb]{ 0,  0,  1}{\textbf{0.63}} & \textcolor[rgb]{ 0,  0,  1}{\textbf{0.63}} \\
	& Qabf ↑ & 0.64  & 0.65  & 0.66  & 0.66  & 0.64  & 0.65  & 0.65  & 0.66  & \textcolor[rgb]{ 0,  0,  1}{\textbf{0.67}} & 0.65  & 0.66  & 0.66  & \textcolor[rgb]{ 0,  0,  1}{\textbf{0.67}} & 0.27  & \textcolor[rgb]{ 1,  0,  0}{\textbf{0.70}} & 0.66 \\
	& Qcb ↑ & 0.67  & 0.68  & \textcolor[rgb]{ 0,  0,  1}{\textbf{0.69}} & \textcolor[rgb]{ 0,  0,  1}{\textbf{0.69}} & \textcolor[rgb]{ 0,  0,  1}{\textbf{0.69}} & \textcolor[rgb]{ 0,  0,  1}{\textbf{0.69}} & 0.68  & \textcolor[rgb]{ 0,  0,  1}{\textbf{0.69}} & \textcolor[rgb]{ 0,  0,  1}{\textbf{0.69}} & \textcolor[rgb]{ 0,  0,  1}{\textbf{0.69}} & \textcolor[rgb]{ 0,  0,  1}{\textbf{0.69}} & \textcolor[rgb]{ 0,  0,  1}{\textbf{0.69}} & \textcolor[rgb]{ 1,  0,  0}{\textbf{0.70}} & 0.66  & \textcolor[rgb]{ 1,  0,  0}{\textbf{0.70}} & \textcolor[rgb]{ 0,  0,  1}{\textbf{0.69}} \\
	& SSIM ↑ & 1.46  & \textcolor[rgb]{ 0,  0,  1}{\textbf{1.48}} & \textcolor[rgb]{ 1,  0,  0}{\textbf{1.49}} & \textcolor[rgb]{ 1,  0,  0}{\textbf{1.49}} & \textcolor[rgb]{ 1,  0,  0}{\textbf{1.49}} & \textcolor[rgb]{ 1,  0,  0}{\textbf{1.49}} & \textcolor[rgb]{ 0,  0,  1}{\textbf{1.48}} & \textcolor[rgb]{ 0,  0,  1}{\textbf{1.48}} & \textcolor[rgb]{ 1,  0,  0}{\textbf{1.49}} & \textcolor[rgb]{ 1,  0,  0}{\textbf{1.49}} & \textcolor[rgb]{ 1,  0,  0}{\textbf{1.49}} & \textcolor[rgb]{ 1,  0,  0}{\textbf{1.49}} & \textcolor[rgb]{ 0,  0,  1}{\textbf{1.48}} & 1.42  & \textcolor[rgb]{ 1,  0,  0}{\textbf{1.49}} & \textcolor[rgb]{ 1,  0,  0}{\textbf{1.49}} \\
	\bottomrule
\end{tabular}%
	\label{tab:ablation}
\end{table*}
\subsection{Performance on Downstream Tasks}We also study the performance of VIS-IR image fusion on two downstream tasks: object detection \textcolor[rgb]{ 0,  0,  0}{and semantic segmentation}.

\label{downstream}
\subsubsection{Performance on Downstream Object Detection}
\label{object}
To assess the object detection performance on source images and fused images, YOLOv5 \cite{yolo}, a SOTA detection network is utilized. We use the M3FD dataset \cite{m3fd}, which has $4200$ pairs of VIS-IR images. The images have objects of six categories: people, cars, trucks, buses, lamps, and motorcycles (M.Cycle). For a fair comparison, we input the visible images, infrared images, and fused images generated using different methods into the YOLOv5 detector for training and testing. We utilize $3800$ images for training and $400$ images for testing. For training, SGD optimizer is used for training $100$ epochs. The batch size and initial learning rate are set to $16$ and $1\times10^{-2}$, respectively. The object detection performance is evaluated using  map@[$0.5:0.95$], which denotes the mean average precision values at different intersection over union (IoU) thresholds (from $0.5$ to $0.95$ in steps of $0.05$). 
Table \ref{tab:object} shows the comparison for object detection performance. \textcolor[rgb]{ 0,  0,  0}{The visible and infrared images have better performance for specific objects. For example, the detector has better performance for detecting people with infrared images, whereas superior car, truck, bus, lamp, and motorcycle detection with visible images. This complementary characteristics can offer better detection performance on fused images.} The results show that except for MDA, all the fusion methods have better detection performance than the source images. Compared to the SOTA fusion methods, our FNet has superior performance.
\textcolor[rgb]{ 0,  0,  0}{\subsubsection{Performance on Downstream Semantic Segmentation}
\label{object}
We utilize SegFormer \cite{segformer}, a SOTA segmentation network, to assess the downstream semantic segmentation performance on source images and fused images. We use $1262$ pairs of VIS-IR images from the MFNet dataset \cite{mfnet}. The images have labels of nine categories: unlabeled (Unlbl), car, person, bike, curve, curve stop (Stop), guardrail (G.Rail), color cone (Cone), and bump. For a fair comparison, we input the visible images, infrared images, and fused images generated using different methods into the SegFormer for training and testing. We utilize $1142$ images for training and $120$ images for testing. AdamW optimizer is used for training $100$ epochs. The batch size and initial learning rate are set to $4$ and $6\times10^{-5}$, respectively. The semantic segmentation performance is evaluated using mean intersection over union (mIoU), which measures the overlap between the predicted segmentation and the ground truth. Table \ref{tab:object} shows the comparison for semantic segmentation performance. The visible and infrared images have better segmentation performance for specific objects, and fusing them improves the scene categorization ability of the segmentation network. 
Except for MDA, all the fusion methods have better segmentation performance than the source images. Compared to the SOTA fusion methods, our FNet has superior performance.}

\textcolor[rgb]{ 0,  0,  0}{The results on downstream tasks show that due to the complementary characteristics of VIS-IR images, the fused images can have better information content and improve the scene analysis performance of downstream networks.}
\subsection{Ablation Study}
\label{ablation}
Our proposed FNet's performance relies on the network design and two-stage training procedure. Especially, our proposed LZSC block effectively estimates the $\ell_0$-regularized common and unique sparse features from the source images, which are then combined to get the final fused image. Moreover, we design an inverse fusion network named IFNet, which is utilized in the training stage I of FNet. In the training stage II, we constrain the fused image to be similar to the source images using a three-component loss function.   
In this section, we conduct ablation experiments (AE) to validate the effectiveness of our proposed method. \textcolor[rgb]{ 0,  0,  0}{Specifically, we study the effectiveness and design choices of our LZSC block and IFNet. We also study the effectiveness of the loss function.} Table \ref{tab:ablation} presents the quantitative comparison of the ablation experiments on eight datasets. Additionally, we show visual results of five ablation experiments for the VIS-IR image fusion task on the RoadScene dataset in Fig. \ref{fig11}.

\noindent
\textbf{Effectiveness of LZSC block:}
Our novel LZSC block is designed to estimate the $\ell_0$-regularized sparse features from an input image. To validate the effectiveness of the LZSC block in FNet and IFNet, we replace it with the LCSC block \cite{lcsc}. The implementation details of the LCSC block is given in Section III-A of the supplementary material. We denote this ablation experiment as \textbf{AE1}, and tabulate the fusion performance in Table \ref{tab:ablation}. 
The visual comparison in Fig. \ref{fig11}-(c) shows that the model with LCSC block produces a fused image with less sharpness.

\noindent
\textcolor[rgb]{ 0,  0,  0}{\textbf{Effects of changing regularization technique in LZSC block:} To validate the effectiveness of $\ell_0$-regularization in the LZSC block against the $\ell_1$-regularization, we conduct ablation experiment \textbf{AE2}. Here, we replace the sigmoidal thresholding
function of the LZSC block in both FNet and IFNet, with the soft thresholding function $S_{\theta}(\cdot)$. As the results show, using $S_{\theta}(\cdot)$ degrades the results in most cases.}

\noindent
\textcolor[rgb]{ 0,  0,  0}{\textbf{Effectiveness of separate convolution layers for different iteration steps in LZSC block: } We use separate convolution layers for different iteration steps in LZSC block. This design is motivated by the work \cite{maximal}, which suggests that having different learnable layers at each iteration helps to improve the restrictive isometry property (RIP) constants of the learned convolutional dictionaries. A dictionary with a smaller value of RIP constant leads to sparse recovery problems that are inherently easier to solve. To validate this design choice, we conduct experiment \textbf{AE3}, where we use shared convolutional layers across iterations. As the result from Table \ref{tab:ablation} show, sharing parameters of convolutional layers across iterations degrade the performance.}

\noindent
\textcolor[rgb]{ 0,  0,  0}{\textbf{Effectiveness of Nesterov momentum in LZSC block:} In our LZSC block, we use Nesterov momentum to improve the convergence speed. To validate this design choice, we conduct three ablation experiments: i) without momentum term (\textbf{AE4}), ii) with Polyak momentum \cite{polyak_classical} (\textbf{AE5}), and iii) with quasi-hyperbolic momentum \cite{qhm} (\textbf{AE6}). We provide implementation details of these three experiments in Section III-B of the supplementary material. As the result from Table \ref{tab:ablation} show, the Nesterov momentum-based LZSC block yielded the best overall performance.}

\noindent
\textbf{Effects of changing $\alpha$ and $\gamma$ in $T_{\alpha,\gamma,\theta}(\cdot)$ of LZSC block:} In our LZSC block, we use the sigmoidal thresholding function $T_{\alpha,\gamma,\theta}(\cdot)$ to achieve $\ell_0$-regularization. As explained in Section \ref{sc}, we use $T_{0.1,100,\theta}(\cdot)$ as a smooth approximation of the hard thresholding function. Decreasing $\alpha$ or increasing $\gamma$ makes the function closer to the hard thresholding function, however it also makes the network training difficult. We conduct ablation experiments to show the effects of increasing $\alpha$ and decreasing $\gamma$. We choose two parameter settings as $\alpha=0.9, \gamma=100$ and $\alpha=0.1, \gamma=50$, and use these values in the LZSC blocks in FNet and IFNet. We denote these two ablation experiments as \textbf{AE7} ($\alpha=0.9, \gamma=100$) and \textbf{AE8} ($\alpha=0.1, \gamma=50$). As shown in Table \ref{tab:ablation}, compared to these two parameter settings, the values used in FNet ($\alpha=0.1, \gamma=100$) gives better results in most cases.

\noindent
\textbf{Effects of changing number of IMs in LZSC block:} We use four iteration modules (IMs) in the LZSC block. To analyze the effects of changing the number of IMs, we use three and five IMs in the LZSC blocks of both FNet and IFNet. \textbf{AE9} (number of IM is 3) and \textbf{AE10} (number of IM is 5) denote the ablation experiments. As evident from the results in Table \ref{tab:ablation}, four IMs used in FNet gives better results in most cases.

\noindent
\textbf{Effectiveness of IFNet:} Our inverse fusion network IFNet is utilized in the training stage-I of FNet. We constrain the source images generated using IFNet to be similar to the original source images, which effectively improves the performance of FNet. To validate the effectiveness of IFNet, we remove the training stage I of FNet. We denote this ablation experiment as \textbf{AE11}. As shown in Fig. \ref{fig11}-(d), w/o IFNet, the model generates blurred fused image. Also, the structure information from source images is not preserved.

\noindent
\textcolor[rgb]{ 0,  0,  0}{\textbf{Effects of changing regularization technique in IFNet:} Our IFNet is a general network architecture, that can be designed with any regularization method. Due to the advantage of $\ell_0$-regularization in nullifying the irrelevant features and preserving the salient features, we design IFNet with $\ell_0$-regularized LZSC block. We conduct an ablation experiment, where we replace the sigmoidal thresholding function of LZSC block with soft thresholding function, to show the effectiveness of $\ell_0$-regularization in IFNet. In this setup, the LZSC blocks in FNet remain unchanged. We denote this experiment as \textbf{AE12}. As evident from the results in Table \ref{tab:ablation}, $\ell_0$-regularization-based design in IFNet gives better results in most cases.}

\noindent
\textbf{Effectiveness of Loss Function:} In the training stage-II, we use a loss function consisting of intensity ($\mathcal{L}_{int}$), gradient ($\mathcal{L}_{grad}$) and ssim ($\mathcal{L}_{ssim}$) loss components. We denote the ablation experiments w/o $\mathcal{L}_{int}$ as \textbf{AE13}, w/o $\mathcal{L}_{grad}$ as \textbf{AE14}. and w/o $\mathcal{L}_{ssim}$ as \textbf{AE15}, respectively. As shown in Fig. \ref{fig11}-(e), w/o $\mathcal{L}_{int}$, the model fails to preserve the texture information. Fig. \ref{fig11}-(f) shows that w/o  $\mathcal{L}_{grad}$, the generated fused images are dark, and the contrast between foreground and background is not maintained. As shown in Fig. \ref{fig11}-(f), w/o $\mathcal{L}_{ssim}$, the model fails to keep the structure information and texture details of the source images.

 The quantitative results in Table \ref{tab:ablation} show that in all the ablation experiments, the MMIF performance degrades to a lesser or greater extent. All these results show the effectiveness of our method.

\section{Conclusion}
\label{sec_5}
This work introduces an interpretable network named FNet for the MMIF task. We design our network based on a novel $\ell_0$-regularized MCSC model, due to which FNet has the advantages of both the interpretability of model-based methods and the efficiency of DNN. Specifically, we propose a novel LZSC block to solve the $\ell_0$-regularized CSC problem. Using three LZSC blocks, FNet separates the source images into unique and common features, which are then combined to get the final fused image. Moreover, we propose a novel $\ell_0$-regularized MCSC model for the inverse fusion process, based on which we design an inverse fusion network named IFNet. Employing IFNet in the training of FNet significantly improves the MMIF performance. Extensive experiments on eight datasets for five MMIF tasks demonstrate that FNet has superior performance compared to the SOTA methods. Besides, FNet also enhances downstream object detection and \textcolor[rgb]{ 0,  0,  0}{semantic segmentation} performance in VIS-IR image pairs. Moreover, we also provide visualization of the unique and common sparse features and reconstruction parts, which shows good interpretability of our network. \textcolor[rgb]{ 0,  0,  0}{Comparison of features with SOTA methods shows that our FNet yields more interpretable features. Additionally, we also demonstrate that $\ell_0$-regularization is implemented in the three LZSC blocks of our proposed FNet.} 
Future work could explore more intricate models to represent the dependency of source images across different modalities for the fusion process and, based on these models, design more interpretable networks for the MMIF task.
\bibliographystyle{IEEEtran}
\bibliography{main}
\begin{IEEEbiography}[{\includegraphics[width=1in,height=1.25in,clip,keepaspectratio]{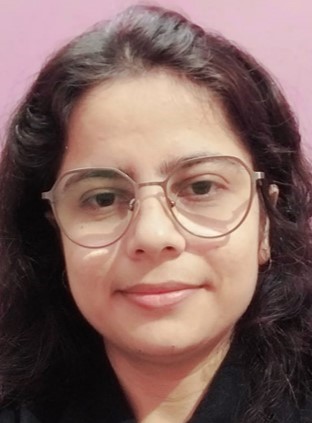}}]{Gargi Panda}
received her M.Tech. degree in electrical engineering from the Indian Institute of Technology (IIT), Kharagpur, India in 2022. She is currently pursuing Ph.D. at the Department of Electrical Engineering, IIT Kharagpur, India. Her research interests include image processing, computer vision, machine learning, and biomedical signal processing.
\end{IEEEbiography}
\begin{IEEEbiography}[{\includegraphics[width=1in,height=1.25in,clip,keepaspectratio]{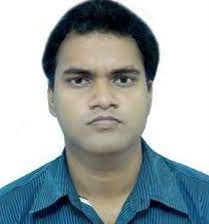}}]{Soumitra Kundu}
received his M.Tech degree in power electronics and electrical drives from the Indian Institute of Technology (Indian School of Mines), IIT(ISM), Dhanbad, India, in 2019. He is currently pursuing Ph.D. at the Rekhi Centre of Excellence for the Science of Happiness, IIT, Kharagpur, India. His research interests include human-machine interfaces (HCI), biomedical signal processing, computer vision, and machine learning.
\end{IEEEbiography}
\begin{IEEEbiography}[{\includegraphics[width=1in,height=1.25in,clip,keepaspectratio]{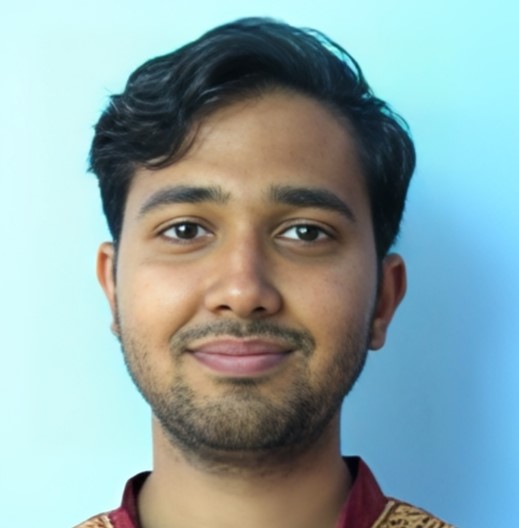}}]{Saumik Bhattacharya} is an assistant professor in the Department of Electronics and Electrical Communication Engineering, Indian Institute of Technology (IIT), Kharagpur, India. He received his Ph.D. degree in Electrical Engineering from the IIT Kanpur, India, in 2017. His research interests include image processing, computer vision, and machine learning.
\end{IEEEbiography}
\begin{IEEEbiography}[{\includegraphics[width=1in,height=1.25in,clip,keepaspectratio]{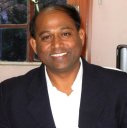}}]{Aurobinda Routray} is a professor in the Department of Electrical Engineering, Indian Institute of Technology (IIT), Kharagpur, India. He received his master's degree in 1991 from IIT, Kanpur, India, and his Ph.D. degree in 1999 from Sambalpur University, Odisha, India. His research interests include big data analytics, computer vision, signal processing, machine learning, cognitive science, and embedded systems.
\end{IEEEbiography}

\end{document}